\documentclass[11pt]{article}

\usepackage[final]{acl}

\usepackage[utf8]{inputenc}
\usepackage[T1]{fontenc}
\usepackage{times}          
\usepackage{latexsym}
\usepackage{microtype}      
\usepackage{inconsolata}    

\usepackage{amsmath}
\usepackage{amssymb}
\usepackage{amsfonts}
\usepackage{bm}
\usepackage{mathtools}

\usepackage[ruled,vlined]{algorithm2e}

\usepackage{graphicx}
\usepackage{hyperref}
\usepackage{booktabs}
\usepackage{multirow}
\usepackage{makecell}
\usepackage[table]{xcolor}  

\usepackage{natbib}

\usepackage{enumitem}
\usepackage{verbatim}
\usepackage{amsthm}
\usepackage{mathrsfs}
\usepackage{subfigure}      

\definecolor{brightblue}{rgb}{0.0, 0.0, 1.0}  

\title{%
  Prompt-R1: Collaborative Automatic Prompting Framework via End-to-end Reinforcement Learning
}

\author{
\bfseries 
Wenjin Liu\textsuperscript{1,2,3} \quad 
Haoran Luo\textsuperscript{2,$\dagger$} \quad 
Xueyuan Lin\textsuperscript{3,4,5} \quad 
Haoming Liu\textsuperscript{6,$\dagger$} \\
\bfseries 
Tiesunlong Shen\textsuperscript{7} \quad 
Jiapu Wang\textsuperscript{8} \quad 
Rui Mao\textsuperscript{2} \quad 
Erik Cambria\textsuperscript{2} \\
\mdseries 
\textsuperscript{1}Hainan University \quad 
\textsuperscript{2}Nanyang Technological University \\
\textsuperscript{3}Hithink Research \quad 
\textsuperscript{4}HKUST (Guangzhou) \quad
\textsuperscript{5}IDEA Research \quad
\textsuperscript{6}Tsinghua University \\
\textsuperscript{7}National University of Singapore \quad 
\textsuperscript{8}Nanjing University of Science and Technology \\
\texttt{wenjinliu23@outlook.com, haoran.luo@ieee.org}\\[0.1em]
\href{https://qwenqking.github.io/Prompt-R1/}{%
\raisebox{-0.2ex}{\includegraphics[height=1em]{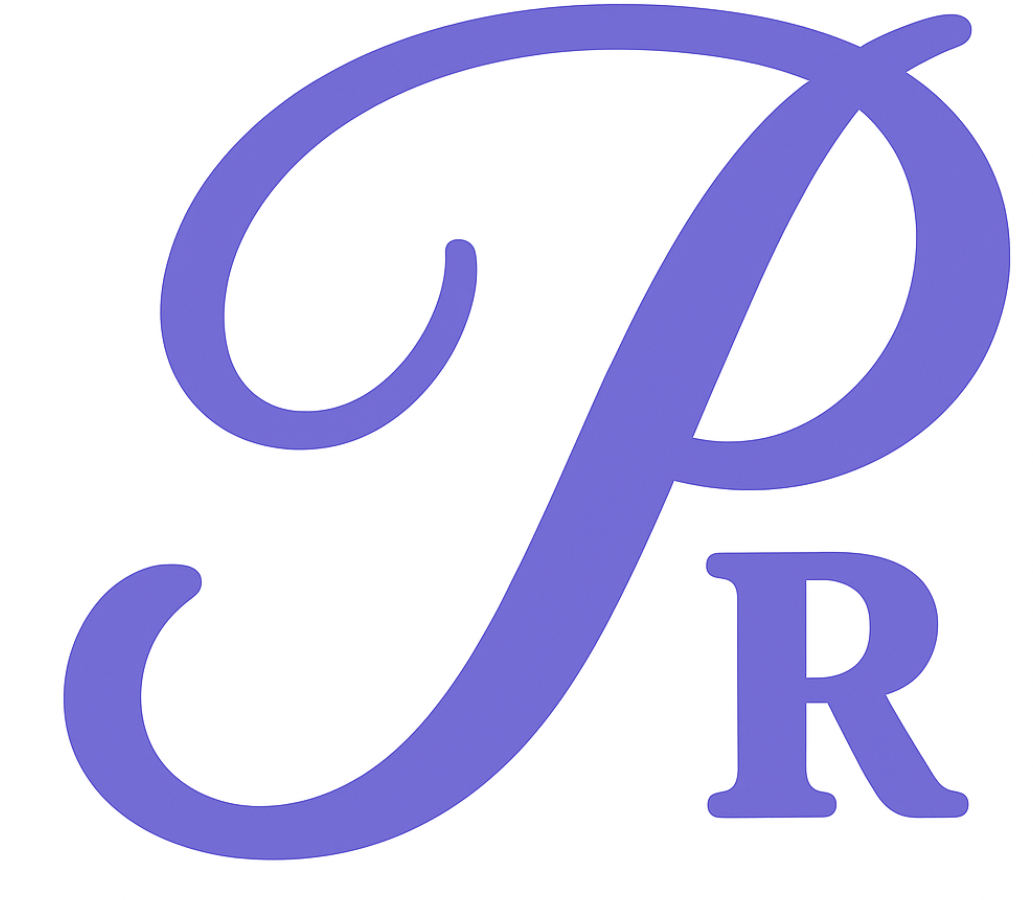}}\;Homepage%
} \quad 
\href{https://github.com/QwenQKing/Prompt-R1}{%
\raisebox{-0.2ex}{\includegraphics[height=1em]{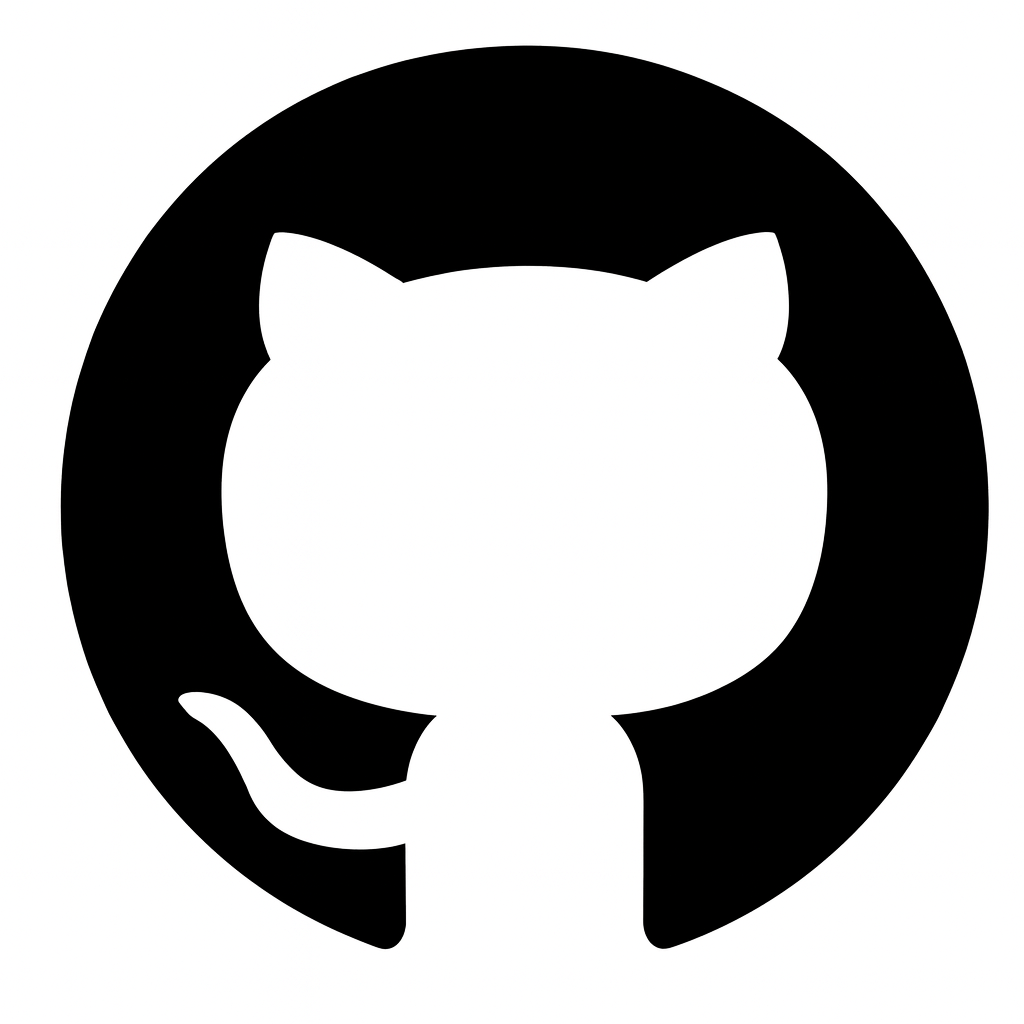}}\;GitHub%
} \quad 
\href{https://huggingface.co/datasets/QwenQKing/Prompt-R1}{%
\raisebox{-0.2ex}{\includegraphics[height=1em]{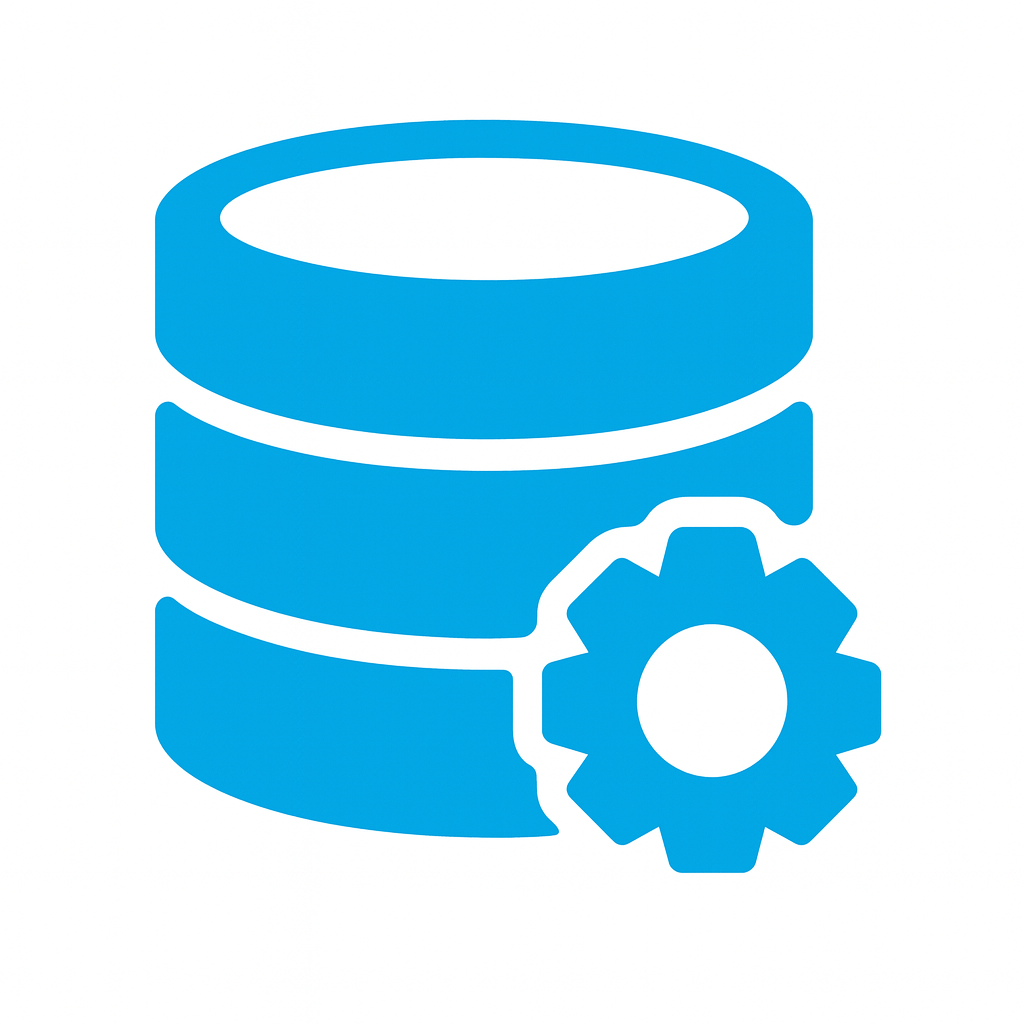}}\;Dataset%
} \quad 
\href{https://huggingface.co/QwenQKing/Prompt-R1}{%
\raisebox{-0.2ex}{\includegraphics[height=1em]{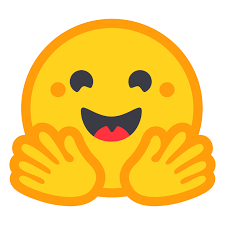}}\;HF Models%
}
}

\begin{document}
\maketitle
\renewcommand{\thefootnote}{\fnsymbol{footnote}}
\footnotetext[0]{$\dagger$ Corresponding authors.}

\begin{abstract}

Recently, various excellent and powerful large language models (LLMs) have been utilized to solve a wide range of human problems. However, when faced with complex problems, most users are often unable to provide accurate and effective prompts to interact with LLMs, thus limiting their performance. To address this challenge, we propose Prompt-R1, an end-to-end reinforcement learning framework that utilizes a small-scale LLM (as \textit{agent}) to collaborate with large-scale LLMs (as \textit{environment}), replacing users to interact better. This collaboration is presented as a multi-turn interaction, where the small-scale LLM thinks and generates prompts, and the large-scale LLM performs complex reasoning. A double-constrained reward is designed to optimize correctness and quality of generation. Prompt-R1 provides a plug-and-play framework that supports both inference and training with various large-scale LLMs. Experimental results on twelve datasets show that Prompt-R1 significantly outperforms baseline LLMs across various tasks.
Our code is available at \url{https://github.com/QwenQKing/Prompt-R1}.
\end{abstract}

\section{Introduction}

In recent years, large language models have been widely applied to assist humans in completing various complex real-world tasks~\citep{wei2025plangenllms,yichao2025stimuli}. Despite their abilities in understanding, reasoning, calculation, and generation, most users fail to fully utilize LLMs, especially in deep reasoning~\citep{zhang2025marsmultiagentadaptivereasoning}, adaptive responses~\citep{yao2024clave}, and multi-turn interaction tasks~\citep{luo2025hypergraphrag,zhang2025ratt}. This is mainly due to users' ineffective construction and adjustment of prompts, limiting LLMs' performance in complex reasoning and dynamic tasks~\citep{do2024prompt,wen2023hard,prakash2025imsorryicant}. Further, the applicability of LLMs to tasks restricts their performance~\citep{ong2025derivingstrategicmarketinsights}. Enhancing LLM's prompt understanding and adaptability is therefore crucial~\citep{zhang2025router}.

\begin{figure}[t]
\vskip 0.2in
\begin{center}
\centerline{\includegraphics[width=1.0\columnwidth]{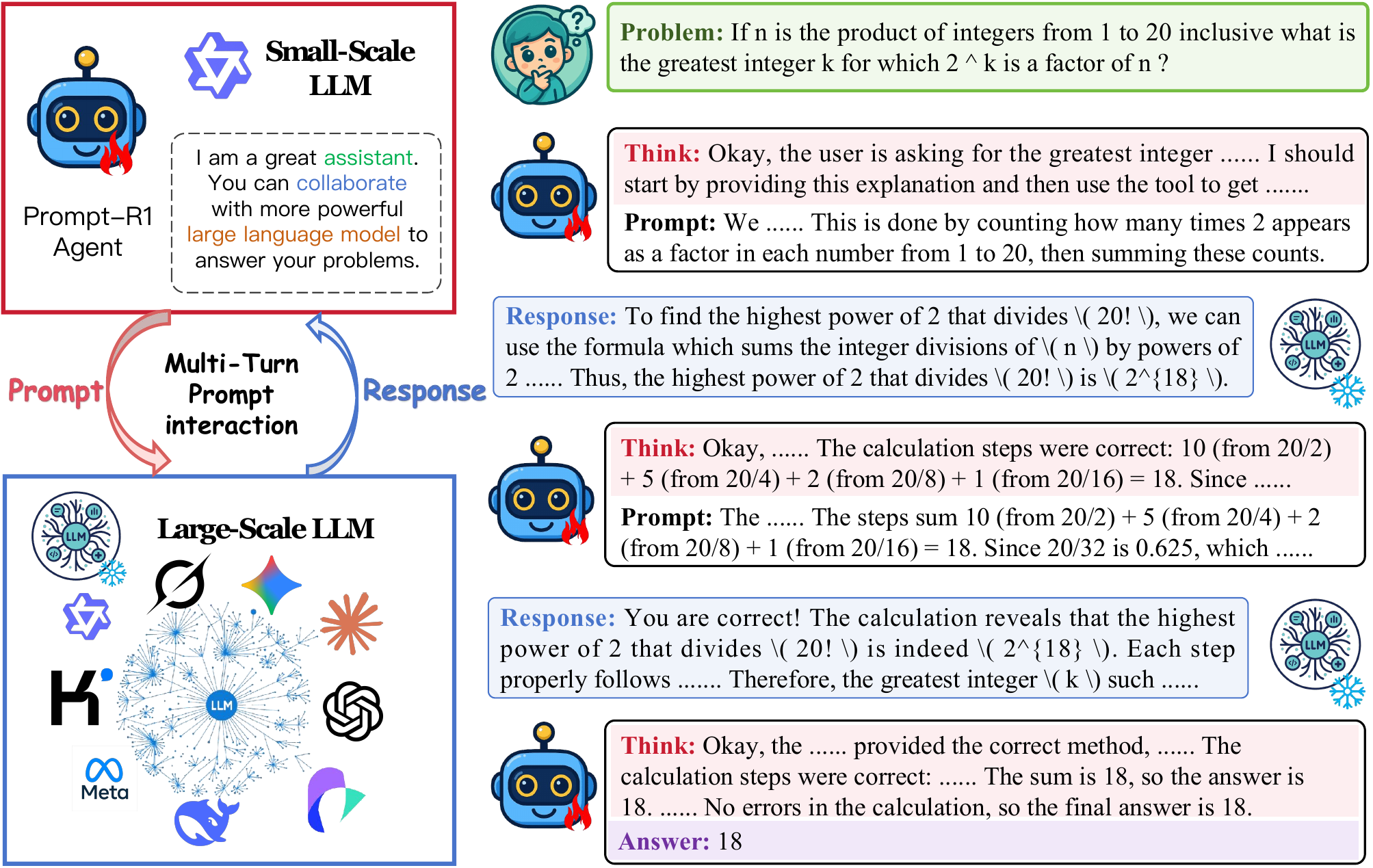}}
\caption{An example of a Prompt-R1 agent working with a large-scale LLM. The agent obtains the correct answer by interacting with the LLM step by step.
}
\label{fig1:inference-example}
\end{center}
\vskip -0.2in
\end{figure}

\begin{figure*}[t]
\centering
\vspace{-0.5em}
\includegraphics[width=1.0\textwidth]{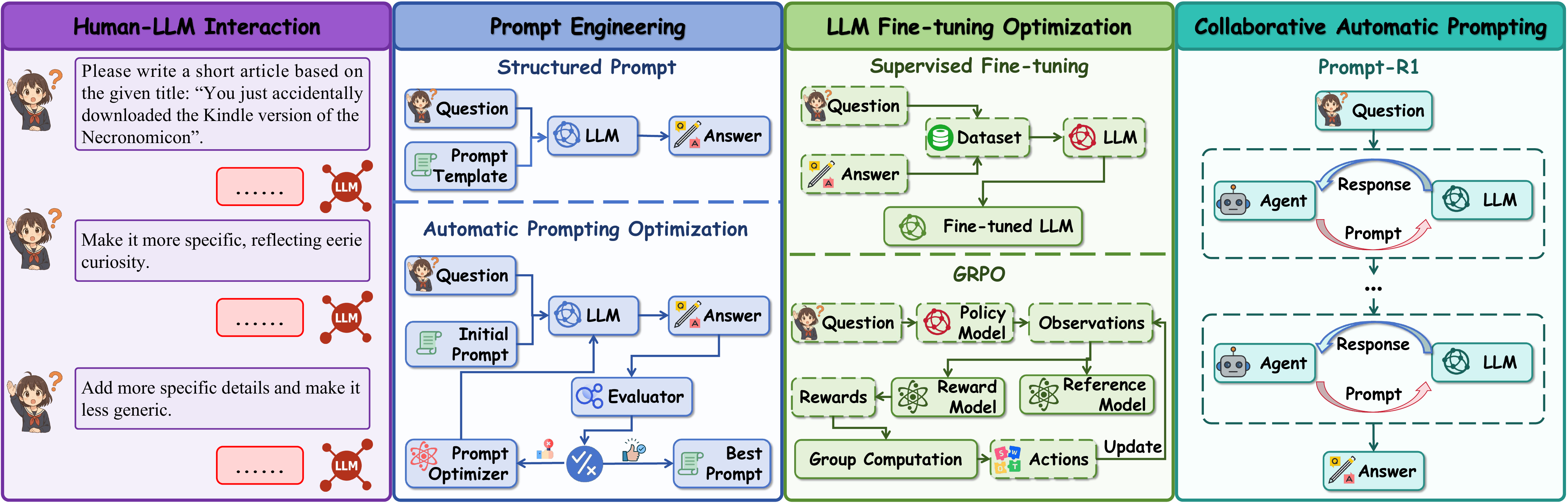}
\caption{Comparison of different methods for improving LLMs' performance: human-LLM interaction, prompt engineering, fine-tuning optimization, and our collaborative automatic prompting interaction framework Prompt-R1.}
\label{fig1:methods_compare}
\end{figure*}

To address these issues, methods based on prompt engineering and fine-tuning have emerged. For prompt engineering-based methods, structured prompting, including chain-of-thought (CoT)~\citep{wei2022chain} and least-to-most prompting~\citep{zhouleast}, and automatic prompt optimization, such as dynamic prompt corruption~\citep{fan2025improving} and TRPrompt~\citep{nica2025trprompt}, enhance LLMs' reasoning, generalization, and adaptation without fine-tuning. In addition, fine-tuning optimization methods such as Low-Rank Adaptation (LoRA)~\citep{hulora}, adapter tuning~\citep{huang2024can}, and instruction tuning~\citep{liao2024instance} improve LLMs' adaptability to task-specific instructions. Furthermore, reinforcement learning-based methods~\citep{luo2025graph} enhance LLMs through continuous feedback, particularly in complex reasoning and multi-turn interaction tasks~\citep{cao2024survey}. These methods significantly improve the prompt understanding and task adaptability for LLMs.

However, these methods still face several challenges:
\textbf{(i) Ability limitations of small-scale LLMs.} Small-scale LLMs have limited understanding and generation capabilities and struggle with long-range dependencies and complex reasoning, limiting their effectiveness in deep comprehension and multi-turn tasks~\citep{luo2025kbqao1}.
\textbf{(ii) High optimization cost for large-scale LLMs.} Fine-tuning large-scale LLMs requires significant computational and storage resources, while official API-based approaches require a large amount of prompt engineering, lack adaptive dynamic optimization, and incur high costs~\citep{wang2025survey}.
\textbf{(iii) Complexity and limited adaptability of large-small-scale LLM collaboration.} Current methods rely on APIs, redundant layers, and cumbersome prompt engineering, increasing costs and reducing collaboration efficiency of small-scale LLMs and large-scale LLMs in dynamic, multi-task environments~\citep{zhang2025collm}.

To address these challenges, we propose Prompt-R1 (see Figure~\ref{fig1:inference-example}), a collaborative automatic prompting framework for the small-scale LLM and large-scale LLM enhanced by end-to-end reinforcement learning (RL)~\citep{guo2025deepseek}, supporting plug-and-play for diverse large-scale LLMs. In this proposed framework, the small-scale LLM acts as an agent through multi-turn prompts interacting with the large-scale LLM as the environment, optimizing prompts and accomplishing tasks better. A double-constrained reward is designed to boost the generation quality and accuracy for the small-scale LLM, while a plug-and-play architecture simplifies coordination and removes API dependency. Prompt-R1 provides a resource-efficient, adaptable, portable, and scalable collaborative paradigm for the small-scale LLM and the large-scale LLM.

We evaluate Prompt-R1 on four tasks: multi-hop reasoning, standard question-answering (QA), mathematical computation, and text generation. Experimental results show Prompt-R1 enhances generation quality and reasoning accuracy through this reinforcement learning-driven multi-turn prompt interaction framework, surpassing baselines and current methods (see Figure~\ref{fig1:methods_compare}). While strengthening the large-scale LLM's reasoning abilities, Prompt-R1 also improves the abilities of small-scale LLMs. Further, it can adapt across tasks without task-specific fine-tuning, demonstrating broad adaptability and strong practical potential.

\section{Related Work}
In this section, we review the current approaches for enhancing the performance of LLMs, including: 

\begin{figure*}[t]
\centering
\vspace{-0.5em}
\includegraphics[width=0.966\textwidth]{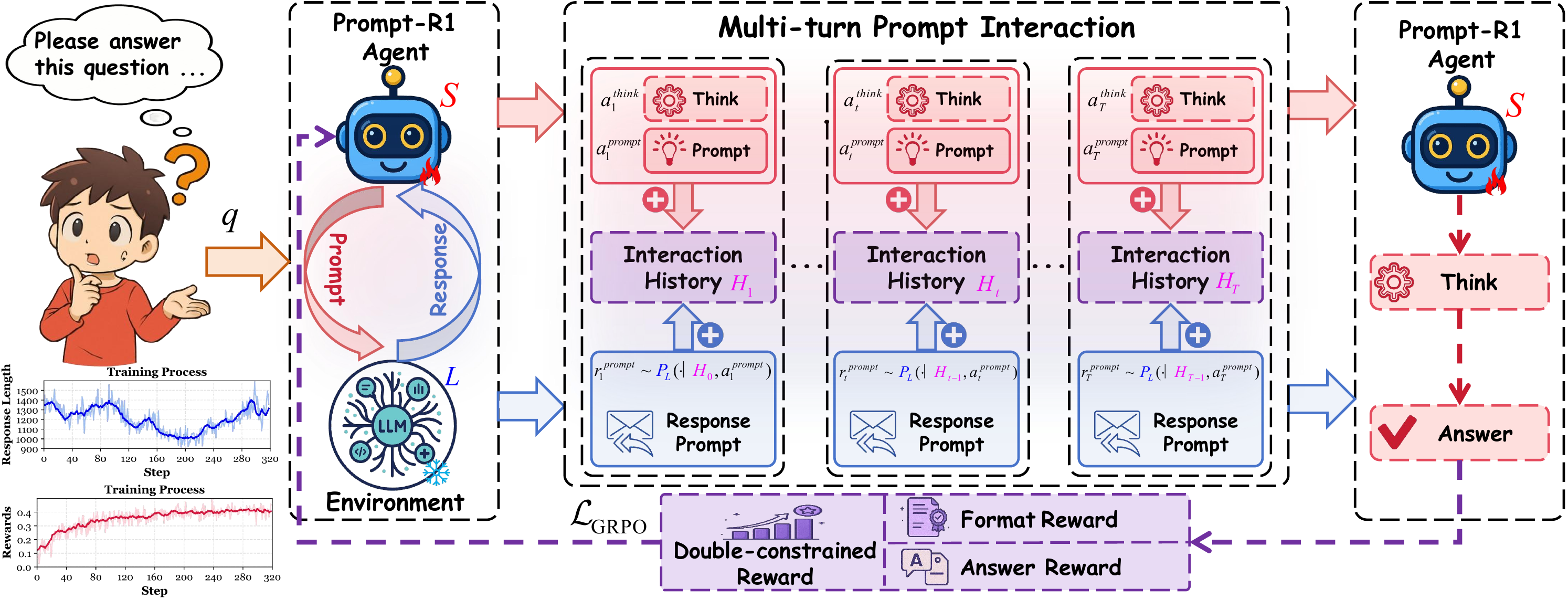} 
\caption{An overview of the Prompt-R1 framework. A small-scale LLM (as agent) interacts with a large-scale LLM (as environment) to answer questions through multi-turn prompts. The large-scale LLM is plug-and-play.}
\label{fig:prompt-r1}
\end{figure*}

\textbf{Automatic Prompting Optimization.} 
Automatic prompt optimization (APO) enhances LLM accuracy and generalization~\citep{liu2025rethinking,zhang2025agentic,asawa2025train}. Prompt Optimization with Textual Gradients~\citep{pryzant2023automatic} utilizes natural-language gradients with beam search and bandits to optimize prompts; TextGrad~\citep{yuksekgonul2024textgrad} applies textual gradients, allowing small LLMs to leverage larger-LLM feedback. Optimization by PROmpting (OPRO)~\citep{yang2023large} treats LLM as an optimizer to refine prompts; Genetic-Pareto (GEPA)~\citep{agrawal2025gepa} mutates candidates and Pareto-selects prompts without weight updates. Residual Optimization Tree \citep{zhou2025riot} adds residual connections to diversify prompts and curb drift, improving its performance. These data- and feedback-driven methods enable exploration without fine-tuning and improve accuracy and scalability~\citep{sunquery,xiao2025prompt}.

\textbf{Reinforcement Learning for LLMs.} 
Reinforcement learning~\citep{Zhang_Wang_Wang_Xu_Lin_Zhang_Mao_Cambria_Liu_2026,tang2025deep} is crucial for improving LLMs in terms of alignment and reasoning~\citep{chaudhari2025rlhf,gu2024review}. Reinforcement Learning with Human Feedback, exemplified by InstructGPT, is foundational~\citep{christiano2017deep,ouyang2022training}. Direct Preference Optimization~\citep{rafailov2023direct}, Odds Ratio Preference Optimization~\citep{hong2024orpo}, and Rank Responses with Human Feedback~\citep{yuan2023rrhf} are single-stage preference optimizers. Reinforcement Learning from artificial intelligence (AI) Feedback~\citep{lee2024rlaif} and Constitutional AI~\citep{bai2022constitutional} lessen reliance on human annotation and improve scalability. Group Relative Policy Optimization (GRPO)~\citep{guo2025deepseek} boosts stability and efficiency, while Agentic RL extends RL to multi-turn and tool-use scenarios, enabling advanced planning and self-correction, as in DeepSeek-R1~\citep{guo2025deepseek}.

\section{Method: Prompt-R1}
In this section, we introduce Prompt-R1 (see Figure \ref{fig:prompt-r1}). It includes Multi-Turn prompt interaction of agent and LLM, double-constrained RL optimization reward, and efficient training and inference.

\subsection{Multi-Turn Interaction of Agent and LLM}
In Prompt-R1, the small-scale and large-scale LLMs collaborate to solve problems through multi-turn interactions. The entire process is as follows:

\textbf{Agent Initialization.}
Prompt-R1 adopts a ReAct-based ~\citep{yao2022react} prompt approach. The initialization of the agent is as follows:

\textit{(i) Environment $L$.}  
The large-scale LLM \(L\) is utilized as the environment in Prompt-R1. The environment \(L\) generates a response message \(r_t^{\mathrm{prompt}}\) based on the current history \(H_{t-1}\) and the collaborative interaction prompt \(a_t^{\mathrm{prompt}}\) from the agent:  
\begin{equation}
r_t^{\mathrm{prompt}} \sim \mathbf{\textcolor{brightblue}{L}}\big(\cdot \mid H_{t-1}, a_t^{\mathrm{prompt}}\big).
\end{equation}
The response \(r_t^{\mathrm{prompt}}\) from the environment \(L\) and the prompt \(a_t^{\mathrm{prompt}}\) from the agent are added to update the multi-turn interaction history \(H_t\):  
\begin{equation}
H_t = H_{t-1} \oplus \big(a_t^{\mathrm{prompt}}, r_t^{\mathrm{prompt}}\big).
\end{equation}
where \(\oplus\) denotes appending the new prompt-response pair to the history. Initially, the history is \(H_0 = [\,]\) and the response is \(r_0^{\mathrm{prompt}} = \varnothing\).

\textit{(ii) Agent \(S\).}  
The small-scale LLM \(S\) acts as the agent. Initially, \(S\) thinks about the question \(q\) and the initial prompt template \(a^{\mathrm{tmpl}}\) (see Table \ref{tab:prompt-tem}), generating the first round of the reasoning process \(a_1^{\mathrm{think}}\) and interaction prompt $a_1^{\mathrm{prompt}}$ for $L$:
\begin{equation}
\big(a_1^{\mathrm{think}}, a_1^{\mathrm{prompt}}\big) \sim  \mathbf{\textcolor{red}{S}}\big(\cdot \mid q, a^{\mathrm{tmpl}}\big).
\end{equation}
Then, the prompt $a_1^{\mathrm{prompt}}$ is sent to $L$ by $S$, and the response $r_1^{\mathrm{prompt}}$ is combined to form the first round history $H_{1}$ for the next round of interaction.

\definecolor{darkred}{rgb}{0,0.6,0}
\definecolor{darkgreen}{rgb}{0, 0.5, 0}
\definecolor{darkpink}{rgb}{0.9,0.1,0.1}
\definecolor{answer}{rgb}{0.2,0,0.5}
\definecolor{question}{rgb}{1.0,0.5,0}
\begin{table*}[t]
\centering
\vspace{-0.5em}
\begin{minipage}{\textwidth}
\hrule height 1pt
\vspace{0.5em}
\fontsize{8.46}{8}\selectfont
First, provide a simple explanation of the question and give it to the large language model for a more accurate answer. Focus on explaining the question without deep reasoning in the first step. After receiving the response, think about the large language model's response, and by interacting with the large language model again and again, arrive at the final answer. Proceed step by step with the following rules:
\texttt{\textbf{\textcolor{darkred}{<think>}}} (don't think deeply and no more than 50 words) \texttt{\textbf{\textcolor{darkred}{</think>}}}
\texttt{\textbf{\textcolor{darkpink}{<interaction\_prompt>}}} (give the question and its explanation to the large language model) \texttt{\textbf{\textcolor{darkpink}{</interaction\_prompt>}}}
After the first step, in each interaction with the large language model, write:
\texttt{\textbf{\textcolor{darkred}{<think>}}} (your reasoning for the receiving response and question) \texttt{\textbf{\textcolor{darkred}{</think>}}}
\texttt{\textbf{\textcolor{darkpink}{<interaction\_prompt>}}} (new request to refine or validate the answer) \texttt{\textbf{\textcolor{darkpink}{</interaction\_prompt>}}}
Each \texttt{\textbf{\textcolor{darkpink}{<interaction\_prompt>}}} must build on what came before. Do not just repeat the same content. Let the content of the \texttt{\textbf{\textcolor{darkpink}{<interaction\_prompt>}}}...\texttt{\textbf{\textcolor{darkpink}{</interaction\_prompt>}}} evolve naturally (for example: outline → add details → refine → check).
Continue producing think within \texttt{\textbf{\textcolor{darkred}{<think>}}}...\texttt{\textbf{\textcolor{darkred}{</think>}}} and call tool within \texttt{\textbf{\textcolor{darkpink}{<interaction\_prompt>}}}...\texttt{\textbf{\textcolor{darkpink}{</interaction\_prompt>}}} until the answer is ready.
Once the answer is complete, write:
\texttt{\textbf{\textcolor{darkred}{<think>}}} (final reasoning with the \texttt{\textbf{\textcolor{brightblue}{<interaction\_response>}}} and question) \texttt{\textbf{\textcolor{darkred}{</think>}}}
\texttt{\textbf{\textcolor{violet}{<answer>}}} (final answer for the question) \texttt{\textbf{\textcolor{violet}{</answer>}}}
\noindent Question: \texttt{\textbf{\textcolor{question}{question}}}.
\vspace{0.36em}
\hrule height 1pt
\end{minipage}
\vspace{-0.3cm}
\caption{The initial prompt template is utilized by the agent $S$ to interact with the environment $L$ (large-scale LLM).}
\label{tab:prompt-tem}
\end{table*}

\textbf{The Agent State Space $\mathcal{H}$.}
The agent state $h_t$ is defined by the multi-turn interaction history of $S$.

\textit{(i) Initial State (\(h_0\)).}  
The initial state of $S$ is \(h_0\), \(h_0 = [\,]\). The state of the first round $h_1$ is based on \(q\) and \(a^{\mathrm{tmpl}}\), as well as composed of $a_1^{\mathrm{think}}$, $a_1^{\mathrm{prompt}}$, and $r_1^{\mathrm{prompt}}$. The first round state \(h_1\) is as follows:  
\begin{equation}
h_1 = [q \oplus a^{\mathrm{tmpl}} \oplus a_1^{\mathrm{think}} \oplus a_1^{\mathrm{prompt}} \oplus r_1^{\mathrm{prompt}}]
\end{equation}

\textit{(ii) State Update (\(h_t\)).}  
From round 2 onward, the state update of $S$ depends on $h_{t-1}$, as well as consists of $a_t^{\mathrm{think}}$, $a_t^{\mathrm{prompt}}$ of $S$, and $r_t^{\mathrm{prompt}}$:
\begin{equation}
h_{t}=[h_{t-1} \oplus a_t^{\mathrm{think}} \oplus a_t^{\mathrm{prompt}} \oplus r_t^{\mathrm{prompt}}].
\end{equation}

\textit{(iii) State Representation (\(F_{h_t}\)).}  
The state representation \( F_{h_t} \) accumulates the complete interaction history up to round \( t \), reflecting all the $a^{\mathrm{think}}$, $a^{\mathrm{prompt}}$, and $r^{\mathrm{prompt}}$. At round \( t \), the state \( F_{h_t} \) is updated by combining the previous state \( F_{h_{t-1}} \), with the current round's reasoning process $a_t^{\text{think}}$, interaction prompt $a_t^{\text{prompt}}$, and the response $r_t^{\text{prompt}}$:
\begin{equation}
F_{h_t} = \textcolor{red}{\mathbf{S}_{t}}(F_{h_{t-1}}, a_t^{\text{think}}, a_t^{\text{prompt}}, r_t^{\text{prompt}} ),
\end{equation}

\textbf{The Agent Action Space.}
The agent \(S\) decides whether to continue reasoning and interacting with \(L\) until the stopping condition is met:
\begin{equation}
\begin{aligned}
\log \mathbf{\textcolor{red}{\pi_\theta}}\big(a_t \mid F_{h_{t-1}}\big)
&= \log \mathbf{\textcolor{red}{\pi_\theta}}\big(a_t^{\mathrm{think}} \mid F_{h_{t-1}}\big) \\
& \hspace{-1.6cm} + \log \mathbf{\textcolor{red}{\pi_\theta}}\big(a_t^{\mathrm{prompt}} \mid F_{h_{t-1}}, a_t^{\mathrm{think}}\big).
\end{aligned}
\end{equation}
During this process, the agent $S$ evaluates its strategy and action probabilities under a stochastic policy, guiding the trajectory toward the final answer.
\textbf{The Agent Target \((h_\ell, F_{h_\ell}, A_{h_\ell})\).}
After multi-turn interactions with the environment $L$, the agent \(S\) will provide the final answer to the question \(q\).

\textit{(i) Final State:}  
The interaction ends at round \(T\), with the final state \(F_{h_T}\) based on the last history \(H_T\), which includes the final response \(r_T^{\mathrm{prompt}}\), provided to $S$ to generate the final answer $A_{h_\ell}$:
\begin{equation}
y = \arg\max_{y \in V^\ast} \mathbf{\textcolor{red}{\pi_\theta}}\big(y \mid q, a^{\mathrm{tmpl}}, H_T\big),
\end{equation}
where \(y\) is the answer \(A_{h_\ell}\) output by the agent \(S\).

\textit{(ii) Final Distribution:}  
The joint distribution of the multi-turn prompt interaction process for the agent \(S\) and the environment \(L\) is as follows:
\begin{equation}
{\fontsize{9.5pt}{10pt}\selectfont
\begin{aligned}
&P_\theta(\tau, y \mid q, a^{\mathrm{tmpl}}) = \\
&\underbrace{\mathbf{\textcolor{red}{\pi_\theta}}(a^{\mathrm{think}}_1, a^{\mathrm{prompt}}_1 \mid q, a^{\mathrm{tmpl}})}_{\text{First round: Prompt generation}} \, 
\underbrace{\mathbf{\textcolor{brightblue}{L}}(r^{\mathrm{prompt}}_1 \mid H_0, a^{\mathrm{prompt}}_1)}_{\text{First round: Response generation}} \\
&\times \prod_{t=2}^{T} \big( \underbrace{\mathbf{\textcolor{red}{\pi_\theta}}(a^{\mathrm{think}}_t, a^{\mathrm{prompt}}_t \mid q, a^{\mathrm{tmpl}}, H_{t-1})}_{\text{Subsequent rounds: Prompt generation}} \big) \\
&\times \big( \underbrace{\mathbf{\textcolor{brightblue}{L}}(r^{\mathrm{prompt}}_t \mid H_{t-1}, a^{\mathrm{prompt}}_t)}_{\text{Subsequent rounds: Response generation}} \big) \\
&\times \underbrace{\mathbf{\textcolor{red}{\pi_\theta^{\mathrm{ans}}}}(a^{\mathrm{think}}_{\mathrm{final}}, y \mid q, a^{\mathrm{tmpl}}, H_T)}_{\text{Final answer}}.
\end{aligned}}
\end{equation}
where $\tau=\big\{(q, a^{\mathrm{tmpl}}, a^{\mathrm{prompt}}_t, r^{\mathrm{prompt}}_t)\big\}_{t=1}^{T}$ is the trajectory for the interactions of the $S$ and $L$; $\pi_\theta$ is the prompt policy of the agent $S$; and $P_L$ is the conditional distribution of the environment $L$.

\par\noindent\textbf{Proposition 1.} \textit{Multi-turn interactions of the small-scale LLM and large-scale LLM can better solve problems.}\vspace{-3mm}
\begin{proof} 
We provide experimental results in sections \ref{4-1} and \ref{4-2}, the case study in Appendix \ref{case_study}, as well as theoretical proofs in Appendix \ref{proof1}.
\end{proof}



\subsection{Double-constrained RL Optimization}

We optimize the agent's policy \( \pi_\theta \) using a dual-constraint reward based on format and correctness, with end-to-end reinforcement learning. The reward \( R \) captures both the format compliance and the answer accuracy at the trajectory level.

\textbf{Double-constrained Reward.} 
To enforce valid reasoning steps and correct answers, we define two components for the output of the agent $S$: the \emph{format reward} $R_{\text{fmt}}$ and the \emph{answer reward} $R_{\text{ans}}$. 

\textit{(i) Format Reward.} 
At round $t$, both reasoning and prompting must be non-empty:
$M_t=\mathbb{I}[a_t^{\mathrm{think}}\neq\varnothing \wedge a_t^{\mathrm{prompt}}\neq\varnothing]$.
At the final turn, we also require a parseable, non-empty answer via $A_p, A_n, C_f$ to ensure completeness and validity:
\begin{equation}
R_{\mathrm{fmt}}=\min\!\big(k,\;\alpha\!\sum_{t=1}^{T-1}\! M_t + \beta A_p + \gamma A_n + \delta C_f\big)
\end{equation}
where \( M_t \) is an indicator ensuring reasoning and prompting are non-empty, preventing incomplete responses; \( A_p \) ensures answer parseability, meaning the output matches the format; \( A_n \) guarantees the non-emptiness of the final answer; \( C_f \) enforces final completeness, ensuring all required components are included; The coefficients \( (\alpha, \beta, \gamma, \delta) \) balance intermediate steps with the goal of a complete answer; and the \( k \) is the upper limit of the format reward, preventing inflation, stabilizing training.

\emph{(ii) Answer Reward.} 
Let $\hat{a}=\mathrm{Norm}(\mathrm{Ans}(y))$ be the normalized predicted answer by the agent $S$, and 
$\mathcal{G}(q)=\{g_i\}$ the reference set. Normalization 
$\mathrm{Norm}(\cdot)$ removes case, punctuation, and articles, while 
$\mathrm{Tok}(\cdot)$ maps text to a multiset of tokens. 
The token-level F1 with a reference $g$ is defined as:
\begin{equation}
\mathrm{F1}(\hat a, g)=\frac{2\,n_{\cap}}{\,|\mathrm{Tok}(\hat a)|+|\mathrm{Tok}(\mathrm{Norm}(g))|\,}.
\end{equation}
where $n_{\cap}$ is the token overlap count between the predicted answer and ground truth. 
The correctness of the predicted answer (answer reward) is:
\begin{equation}
R_{\text{ans}} = \max_{g \in \mathcal{G}(q)} \mathrm{F1}(\hat{a}, g).
\end{equation}
\emph{(iii) Gated Composition for Double-constrained Reward.} 
The overall reward $R$ includes the format reward $R_{\text{fmt}}$ and the answer correctness reward $R_{\text{ans}}$. The calculation of the overall reward $R$ is:
\begin{equation}
R = 
\begin{cases}
-k + R_{\text{fmt}} + R_{\text{ans}}, & R_{\text{fmt}}=k,\\
-k + R_{\text{fmt}}, & \text{otherwise}.
\end{cases}
\end{equation}
so that the correctness for the output answer $\pi_\theta$ of the agent is only credited when the format conditions are fully and correctly satisfied.

\textbf{End-to-End Reinforcement Learning.} 
We adopt a GRPO-based objective, standardizing rewards in a batch of \(M\) trajectories. Let \(R^{(i)}\) be the reward of trajectory \(i\), with the mean reward \(\bar{R}\):
\begin{equation}
\hat{A}^{(i)} = \frac{R^{(i)} - \bar{R}}{\sqrt{\tfrac{1}{M}\sum_{j=1}^M (R^{(j)}-\bar{R})^2 + \varepsilon}},
\end{equation}
where $\hat{A}^{(i)}$ is the standardized advantage, and $\varepsilon$ is a stability constant.
The GRPO-based objective is:
\begin{equation}
{\fontsize{10pt}{10pt}\selectfont
\begin{aligned}
J_{\text{GRPO}}(\theta) &= \mathbb{E}_{\tau \sim p_{\textcolor{red}{\theta_S}, \textcolor{blue}{\theta_L}}(\tau)} \big[ \frac{1}{M} \sum_{i=1}^M \big( \frac{1}{|\tau^{(i)}|}\sum_{t=1}^{|\tau^{(i)}|} \min \big( \\
&\hspace{-1.3cm} \frac{\textcolor{red}{\pi_\theta}(w_t^{(i)} \mid \tau_{<t}^{(i)})}{{\pi_{\theta_{\text{old}}}}(w_t^{(i)} \mid \tau_{<t}^{(i)})} \hat{A}(\tau^{(i)}), \text{clip}\!\big(\frac{\textcolor{red}{\pi_\theta}(w_t^{(i)} \mid \tau_{<t}^{(i)})}{\pi_{\theta_{\text{old}}}(w_t^{(i)} \mid \tau_{<t}^{(i)})},1 \pm \epsilon \big)\\
& \hspace{-1.3cm} \times \hat{A}(\tau^{(i)}) \big) - \beta_{kl} \mathbb{D}_{\text{KL}}(\textcolor{red}{\pi_\theta} \parallel \pi_{\text{ref}}) \big) \big],
\end{aligned}}
\end{equation}
where $p_{\theta_S, \theta_L}(\tau)$ is the joint distribution of $S$ and $L$; $w_t^{(i)}$ is the $t$-th token of $\tau^{(i)}$; and $\pi_{\theta_{\text{old}}}$ and $\pi_{\text{ref}}$ are the pre-update and reference policies, respectively. The $\text{clip}(\cdot)$ limits policy ratios to $1 \pm \epsilon$ to stabilize updates. A KL term $\mathbb{D}_{\text{KL}}({\pi_\theta} \parallel \pi_{\text{ref}})$ regularizes $\pi_{\text{ref}}$, with $\beta_{kl}$ controlling its strength.

\par\noindent\textbf{Proposition 2.} \textit{Reinforcement learning can make small-scale LLMs better guide large-scale LLMs to complete tasks.}\vspace{-3mm}
\begin{proof} 
We provide experimental results in Section \ref{4-4} and Section \ref{4-5} and theoretical proofs in Appendix \ref{proof2}.
\end{proof}

\newcommand{\coloropacity}{6}  

\definecolor{traincolor}{RGB}{42, 170, 138}   
\definecolor{avgcolor}{RGB}{0, 150, 255}      

\newcommand{\increase}[2]{#1 {\color{green!60!black}(+#2)}}
\newcommand{\decrease}[2]{#1 {\color{red}(-#2)}}

\begin{table*}[t]

\vskip -0.15in
\begin{center}
\begin{small}
\fontsize{8pt}{8pt}\selectfont
\setlength{\tabcolsep}{1.86pt}
\setlength{\heavyrulewidth}{1.2pt}
\setlength{\lightrulewidth}{0.36pt}
\setlength{\arrayrulewidth}{0.36pt}
\begin{tabular}{l l c c c c c c c c c c}
\toprule
\textbf{Dataset} & \textbf{Metric} &
\multicolumn{2}{c}{\textbf{Baseline}} &
\multicolumn{1}{c}{\textbf{SFT}} &
\multicolumn{2}{c}{\textbf{CoT Reasoning}} &
\multicolumn{1}{c}{\textbf{GRPO}} &
\multicolumn{3}{c}{\textbf{APO (GPT-4o-mini)}} &
\multicolumn{1}{c}{\textbf{Ours}} \\
\cmidrule(lr){3-4}\cmidrule(lr){5-5}\cmidrule(lr){6-7}\cmidrule(lr){8-8}\cmidrule(lr){9-11}\cmidrule(lr){12-12}
& &
\makecell{Qwen3-4B} & \makecell{GPT-4o-mini} &
\makecell{Qwen3-4B} &
\makecell{Qwen3-4B} & \makecell{GPT-4o-mini} &
\makecell{Qwen3-4B} &
OPRO & TextGrad & GEPA & \makecell{\textbf{Prompt-R1} \textbf{($\Delta
\uparrow$)}}\\
\midrule
\multirow{2}{*}{\textbf{2Wiki}} & \cellcolor{traincolor!\coloropacity}\textbf{EM} & \cellcolor{traincolor!\coloropacity}28.13 & \cellcolor{traincolor!\coloropacity}33.59 & \cellcolor{traincolor!\coloropacity}41.41 & \cellcolor{traincolor!\coloropacity}21.88 & \cellcolor{traincolor!\coloropacity}43.75 & \cellcolor{traincolor!\coloropacity}34.38 & \cellcolor{traincolor!\coloropacity}25.00 & \cellcolor{traincolor!\coloropacity}18.75 & \cellcolor{traincolor!\coloropacity}41.41 & \cellcolor{traincolor!\coloropacity}\textbf{\increase{48.44}{14.85}} \\
 & \cellcolor{traincolor!\coloropacity}\textbf{F1} & \cellcolor{traincolor!\coloropacity}29.32 & \cellcolor{traincolor!\coloropacity}36.57 & \cellcolor{traincolor!\coloropacity}42.62 & \cellcolor{traincolor!\coloropacity}24.17 & \cellcolor{traincolor!\coloropacity}49.13 & \cellcolor{traincolor!\coloropacity}35.05 & \cellcolor{traincolor!\coloropacity}35.96 & \cellcolor{traincolor!\coloropacity}27.50 & \cellcolor{traincolor!\coloropacity}46.27 & \cellcolor{traincolor!\coloropacity}\textbf{\increase{54.41}{17.84}} \\
\midrule
\multirow{2}{*}{\textbf{Hotpot}} & \cellcolor{traincolor!\coloropacity}\textbf{EM} & \cellcolor{traincolor!\coloropacity}21.09 & \cellcolor{traincolor!\coloropacity}33.59 & \cellcolor{traincolor!\coloropacity}23.44 & \cellcolor{traincolor!\coloropacity}18.75 & \cellcolor{traincolor!\coloropacity}42.97 & \cellcolor{traincolor!\coloropacity}27.34 & \cellcolor{traincolor!\coloropacity}34.38 & \cellcolor{traincolor!\coloropacity}27.34 & \cellcolor{traincolor!\coloropacity}38.28 & \cellcolor{traincolor!\coloropacity}\textbf{\increase{44.53}{10.94}} \\
 & \cellcolor{traincolor!\coloropacity}\textbf{F1} & \cellcolor{traincolor!\coloropacity}24.25 & \cellcolor{traincolor!\coloropacity}40.07 & \cellcolor{traincolor!\coloropacity}31.09 & \cellcolor{traincolor!\coloropacity}22.98 & \cellcolor{traincolor!\coloropacity}49.70 & \cellcolor{traincolor!\coloropacity}32.27 & \cellcolor{traincolor!\coloropacity}46.83 & \cellcolor{traincolor!\coloropacity}37.10 & \cellcolor{traincolor!\coloropacity}47.03 & \cellcolor{traincolor!\coloropacity}\textbf{\increase{52.31}{12.24}} \\
\midrule
\multirow{2}{*}{\textbf{GSM8K}} & \cellcolor{traincolor!\coloropacity}\textbf{EM} & \cellcolor{traincolor!\coloropacity}84.38 & \cellcolor{traincolor!\coloropacity}83.59 & \cellcolor{traincolor!\coloropacity}32.03 & \cellcolor{traincolor!\coloropacity}82.81 & \cellcolor{traincolor!\coloropacity}84.38 & \cellcolor{traincolor!\coloropacity}92.97 & \cellcolor{traincolor!\coloropacity}63.28 & \cellcolor{traincolor!\coloropacity}70.31 & \cellcolor{traincolor!\coloropacity}87.50 & \cellcolor{traincolor!\coloropacity}\textbf{\increase{97.66}{14.07}} \\
 & \cellcolor{traincolor!\coloropacity}\textbf{F1} & \cellcolor{traincolor!\coloropacity}84.38 & \cellcolor{traincolor!\coloropacity}86.72 & \cellcolor{traincolor!\coloropacity}32.03 & \cellcolor{traincolor!\coloropacity}82.81 & \cellcolor{traincolor!\coloropacity}88.02 & \cellcolor{traincolor!\coloropacity}92.97 & \cellcolor{traincolor!\coloropacity}83.65 & \cellcolor{traincolor!\coloropacity}85.99 & \cellcolor{traincolor!\coloropacity}90.10 & \cellcolor{traincolor!\coloropacity}\textbf{\increase{97.66}{10.94}} \\
\midrule
\multirow{2}{*}{\textbf{DAPO}} & \cellcolor{traincolor!\coloropacity}\textbf{EM} & \cellcolor{traincolor!\coloropacity}0.00 & \cellcolor{traincolor!\coloropacity}18.75 & \cellcolor{traincolor!\coloropacity}3.13 & \cellcolor{traincolor!\coloropacity}0.00 & \cellcolor{traincolor!\coloropacity}20.31 & \cellcolor{traincolor!\coloropacity}3.91 & \cellcolor{traincolor!\coloropacity}6.25 & \cellcolor{traincolor!\coloropacity}10.16 & \cellcolor{traincolor!\coloropacity}13.28 & \cellcolor{traincolor!\coloropacity}\textbf{\increase{26.56}{7.81}} \\
 & \cellcolor{traincolor!\coloropacity}\textbf{F1} & \cellcolor{traincolor!\coloropacity}0.00 & \cellcolor{traincolor!\coloropacity}18.76 & \cellcolor{traincolor!\coloropacity}3.13 & \cellcolor{traincolor!\coloropacity}0.00 & \cellcolor{traincolor!\coloropacity}20.32 & \cellcolor{traincolor!\coloropacity}3.91 & \cellcolor{traincolor!\coloropacity}6.39 & \cellcolor{traincolor!\coloropacity}10.27 & \cellcolor{traincolor!\coloropacity}14.06 & \cellcolor{traincolor!\coloropacity}\textbf{\increase{26.56}{7.80}} \\
\midrule
\multirow{2}{*}{\textbf{MusiQue}} & \cellcolor{traincolor!\coloropacity}\textbf{EM} & \cellcolor{traincolor!\coloropacity}1.56 & \cellcolor{traincolor!\coloropacity}14.06 & \cellcolor{traincolor!\coloropacity}7.81 & \cellcolor{traincolor!\coloropacity}3.13 & \cellcolor{traincolor!\coloropacity}17.97 & \cellcolor{traincolor!\coloropacity}8.59 & \cellcolor{traincolor!\coloropacity}14.06 & \cellcolor{traincolor!\coloropacity}14.84 & \cellcolor{traincolor!\coloropacity}15.63 & \cellcolor{traincolor!\coloropacity}\textbf{\increase{18.75}{4.69}} \\
 & \cellcolor{traincolor!\coloropacity}\textbf{F1} & \cellcolor{traincolor!\coloropacity}5.44 & \cellcolor{traincolor!\coloropacity}22.06 & \cellcolor{traincolor!\coloropacity}16.78 & \cellcolor{traincolor!\coloropacity}6.95 & \cellcolor{traincolor!\coloropacity}25.39 & \cellcolor{traincolor!\coloropacity}13.90 & \cellcolor{traincolor!\coloropacity}26.18 & \cellcolor{traincolor!\coloropacity}24.06 & \cellcolor{traincolor!\coloropacity}24.91 & \cellcolor{traincolor!\coloropacity}\textbf{\increase{26.31}{4.25}} \\
\midrule
\multirow{2}{*}{\textbf{PopQA}} & \cellcolor{traincolor!\coloropacity}\textbf{EM} & \cellcolor{traincolor!\coloropacity}7.03 & \cellcolor{traincolor!\coloropacity}25.78 & \cellcolor{traincolor!\coloropacity}7.81 & \cellcolor{traincolor!\coloropacity}7.03 & \cellcolor{traincolor!\coloropacity}28.13 & \cellcolor{traincolor!\coloropacity}7.81 & \cellcolor{traincolor!\coloropacity}23.44 & \cellcolor{traincolor!\coloropacity}19.53 & \cellcolor{traincolor!\coloropacity}27.34 & \cellcolor{traincolor!\coloropacity}\textbf{\increase{28.13}{2.35}} \\
 & \cellcolor{traincolor!\coloropacity}\textbf{F1} & \cellcolor{traincolor!\coloropacity}9.97 & \cellcolor{traincolor!\coloropacity}30.75 & \cellcolor{traincolor!\coloropacity}9.53 & \cellcolor{traincolor!\coloropacity}9.58 & \cellcolor{traincolor!\coloropacity}32.66 & \cellcolor{traincolor!\coloropacity}12.27 & \cellcolor{traincolor!\coloropacity}30.14 & \cellcolor{traincolor!\coloropacity}28.28 & \cellcolor{traincolor!\coloropacity}31.18 & \cellcolor{traincolor!\coloropacity}\textbf{\increase{33.77}{3.02}} \\
\midrule
\multirow{2}{*}{\textbf{BookSum}} & \cellcolor{traincolor!\coloropacity}\textbf{F1} & \cellcolor{traincolor!\coloropacity}19.99 & \cellcolor{traincolor!\coloropacity}20.45 & \cellcolor{traincolor!\coloropacity}25.47 & \cellcolor{traincolor!\coloropacity}16.68 & \cellcolor{traincolor!\coloropacity}11.58 & \cellcolor{traincolor!\coloropacity}25.88 & \cellcolor{traincolor!\coloropacity}16.18 & \cellcolor{traincolor!\coloropacity}25.79 & \cellcolor{traincolor!\coloropacity}0.00 & \cellcolor{traincolor!\coloropacity}\textbf{\increase{26.50}{6.05}} \\
 & \cellcolor{traincolor!\coloropacity}\textbf{SSim} & \cellcolor{traincolor!\coloropacity}45.19 & \cellcolor{traincolor!\coloropacity}57.01 & \cellcolor{traincolor!\coloropacity}54.55 & \cellcolor{traincolor!\coloropacity}35.02 & \cellcolor{traincolor!\coloropacity}56.51 & \cellcolor{traincolor!\coloropacity}54.21 & \cellcolor{traincolor!\coloropacity}23.22 & \cellcolor{traincolor!\coloropacity}34.95 & \cellcolor{traincolor!\coloropacity}0.00 & \cellcolor{traincolor!\coloropacity}\textbf{\increase{60.87}{3.86}} \\
\midrule
\multirow{2}{*}{\textbf{W.P.}} & \cellcolor{traincolor!\coloropacity}\textbf{F1} & \cellcolor{traincolor!\coloropacity}13.06 & \cellcolor{traincolor!\coloropacity}19.48 & \cellcolor{traincolor!\coloropacity}15.12 & \cellcolor{traincolor!\coloropacity}10.20 & \cellcolor{traincolor!\coloropacity}9.93 & \cellcolor{traincolor!\coloropacity}8.40 & \cellcolor{traincolor!\coloropacity}7.81 & \cellcolor{traincolor!\coloropacity}22.07 & \cellcolor{traincolor!\coloropacity}0.21 & \cellcolor{traincolor!\coloropacity}\textbf{\increase{22.11}{2.63}} \\
 & \cellcolor{traincolor!\coloropacity}\textbf{SSim} & \cellcolor{traincolor!\coloropacity}20.59 & \cellcolor{traincolor!\coloropacity}35.31 & \cellcolor{traincolor!\coloropacity}30.86 & \cellcolor{traincolor!\coloropacity}12.15 & \cellcolor{traincolor!\coloropacity}30.86 & \cellcolor{traincolor!\coloropacity}10.41 & \cellcolor{traincolor!\coloropacity}13.73 & \cellcolor{traincolor!\coloropacity}20.28 & \cellcolor{traincolor!\coloropacity}0.34 & \cellcolor{traincolor!\coloropacity}\textbf{\increase{38.54}{3.23}} \\
\midrule
\multirow{3}{*}{\textbf{Average}} & \cellcolor{avgcolor!\coloropacity}\textbf{EM}   & \cellcolor{avgcolor!\coloropacity}23.70 & \cellcolor{avgcolor!\coloropacity}34.89 & \cellcolor{avgcolor!\coloropacity}19.27 & \cellcolor{avgcolor!\coloropacity}22.27 & \cellcolor{avgcolor!\coloropacity}39.59 & \cellcolor{avgcolor!\coloropacity}29.17 & \cellcolor{avgcolor!\coloropacity}27.74 & \cellcolor{avgcolor!\coloropacity}26.82 & \cellcolor{avgcolor!\coloropacity}37.24 & \cellcolor{avgcolor!\coloropacity}\textbf{\increase{44.01}{9.12}} \\
 & \cellcolor{avgcolor!\coloropacity}\textbf{F1}   & \cellcolor{avgcolor!\coloropacity}23.30 & \cellcolor{avgcolor!\coloropacity}34.36 & \cellcolor{avgcolor!\coloropacity}21.97 & \cellcolor{avgcolor!\coloropacity}21.67 & \cellcolor{avgcolor!\coloropacity}35.84 & \cellcolor{avgcolor!\coloropacity}28.08 & \cellcolor{avgcolor!\coloropacity}31.64 & \cellcolor{avgcolor!\coloropacity}32.63 & \cellcolor{avgcolor!\coloropacity}31.72 & \cellcolor{avgcolor!\coloropacity}\textbf{\increase{42.45}{8.09}} \\
 & \cellcolor{avgcolor!\coloropacity}\textbf{SSim} & \cellcolor{avgcolor!\coloropacity}32.89 & \cellcolor{avgcolor!\coloropacity}46.16 & \cellcolor{avgcolor!\coloropacity}42.71 & \cellcolor{avgcolor!\coloropacity}23.59 & \cellcolor{avgcolor!\coloropacity}43.69 & \cellcolor{avgcolor!\coloropacity}32.31 & \cellcolor{avgcolor!\coloropacity}18.48 & \cellcolor{avgcolor!\coloropacity}27.62 & \cellcolor{avgcolor!\coloropacity}0.17  & \cellcolor{avgcolor!\coloropacity}\textbf{\increase{49.71}{3.55}} \\
\bottomrule
\end{tabular}
\end{small}
\end{center}
\vskip -0.1in
\caption{Comparison of the selected state-of-the-art baselines and the proposed Prompt-R1 on four tasks, which consists of \textbf{multi-hop reasoning} (\textbf{2Wiki}: 2WikiMultihopQA, \textbf{Hotpot}: HotpotQA), \textbf{mathematical computation} (\textbf{GSM8K}, \textbf{DAPO}: DAPO Math), \textbf{standard QA} (\textbf{MusiQue}, \textbf{PopQA}), and \textbf{text generation} (\textbf{BookSum}, \textbf{W.P.}: WritingPrompts). $\Delta\uparrow$ is the gap between the Prompt-R1 and the baseline large language model (GPT-4o-mini), where the higher values indicating better performance. Bold values are the best performance. All values are in \%. }
\label{tab:prompt-r1-train}
\end{table*}



\subsection{Efficient Training and Inference}
Let the set of available large-scale LLMs (\textit{environment}) be $\mathcal{M} = \{m_1, \dots, m_K\}$, and let $\mathcal{V}^*$ represent the space of finite-length token sequences. The user question $q \in \mathcal{V}^*$, the agent $S$ adopts the policy $\pi_\theta(p_t | q, a^{\mathrm{tmpl}}, H_{t-1})$ to produce the turn-$t$ collaborative prompt $p_t$, where $H_{t-1}$ is the previous interaction history. The multi-turn prompt interaction history $H_t = \{(p_1, r_1), \dots, (p_{t-1}, r_{t-1})\}$ is updated as $H_t = H_{t-1} \oplus (p_t, r_{t})$, with $p_t, r_t \in \mathcal{V}^*$ as the current prompt-response pair, and the interaction history of the agent $S$ and the environment $L$ is updated accordingly, progressively adapting.

\textbf{Training.}
Choose an environment \(m_{\text{train}} \in \mathcal{M}\). The interaction and history update are as follows:
\begin{equation}
\begin{gathered}
p_t \sim \textcolor{red}{\pi_\theta}( \cdot \mid q, a^{\mathrm{tmpl}}, H_{t-1}),\\
r_t \sim \textcolor{brightblue}{P_L^{(m_{\text{train}})}}( \cdot \mid H_{t-1}, p_t),\\
H_t = H_{t-1} \oplus (p_t, r_t), \\
\end{gathered}
\end{equation}
where \(T\) is the number of interaction turns; after $T$-turns, the agent $S$ generates the final answer $y$.

\textbf{Inference.}
Choose an environment \(m_{\text{test}} \in \mathcal{M}\); 
optionally define a session-level routing function \(\rho: \mathcal{V}^* \times (\mathcal{V}^*)^* \to 
\Delta(\mathcal{M})\) and set:
\begin{equation}
\begin{gathered}
p_t \sim \textcolor{red}{\pi_\theta}( \cdot \mid q, a^{\mathrm{tmpl}}, H_{t-1}),  \\
r_t \sim \textcolor{brightblue}{P_L^{(m_{\text{test}})}}( \cdot \mid H_{t-1}, p_t), \\
H_t = H_{t-1} \oplus (p_t, r_t),\\
\end{gathered}
\end{equation}
where \(\Delta(\mathcal{M})\) is the probability simplex over \(\mathcal{M}\), and \(\rho(m | q)\) is the probability of selecting a testing LLM based on the user's question \(q\). Therefore, Prompt-R1 unifies heterogeneous LLMs by training with \(m_{\text{train}}\) to learn the policy \(\pi_\theta\) and performing inference with \(m_{\text{test}}\), ensuring compatibility with a variety of LLMs without altering \(\pi_\theta\).

\par\noindent\textbf{Proposition 3.} \textit{The agent can enhance not only the LLM used for its training but also other LLMs.}\vspace{-3mm}
\begin{proof} 
We provide experimental results in Section \ref{exp-proof3} and theoretical proofs in Appendix \ref{proof3}.
\end{proof}

\section{Experiments}
In this section, we present the experimental setup and results. We address the following research questions (RQs): RQ1: Does Prompt-R1 outperform other methods? RQ2: How is the generalization ability of Prompt-R1? RQ3: How is the transferability of Prompt-R1? RQ4: Does the main component of Prompt-R1 work effectively? RQ5: How do the different environments (zero-cost and overhead-cost) impact Prompt-R1's performance?

\subsection{Experimental Setup}
\textbf{Datasets.}  
To assess the performance of Prompt-R1, we conducted training and evaluation on eight datasets and generalized testing on four out-of-distribution (OOD) datasets. The twelve datasets include \textbf{2WikiMultihopQA}~\citep{ho2020constructing}, \textbf{HotpotQA}~\citep{yang2018hotpotqa}, \textbf{GSM8K}~\citep{cobbe2021training}, \textbf{DAPO Math 17K}~\citep{yu2025dapo}, \textbf{MusiQue}~\citep{trivedi2022musique}, \textbf{PopQA}~\citep{mallen2023not}, \textbf{BookSum}~\citep{kryscinski2022booksum}, and \textbf{WritingPrompts}~\citep{huang2024gpt}. For generalization evaluation, we used \textbf{MathQA}~\citep{amini2019mathqa}, \textbf{SQuAD v2}~\citep{rajpurkar2018know}, \textbf{TriviaQA}~\citep{joshi2017triviaqa}, and \textbf{XSum}~\citep{narayan2018don}. More details of the datasets are illustrated in Appendix \ref{dataset}.

\textbf{Baselines.}
We compare the Prompt-R1 with several baselines and state-of-the-art (SOTA) methods, including \textbf{SFT} (Qwen3-4B), \textbf{CoT Reasoning} (Qwen3-4B and GPT-4o-mini), \textbf{GRPO} (Qwen3-4B), and \textbf{APO} methods (e.g., OPRO, TextGrad, and GEPA). More details are in Appendix \ref{baselines}.

\textbf{Evaluation Metrics.}
In the experiments, we employed three evaluation metrics for Prompt-R1, including Exact Match (\textbf{EM}), F1 score (\textbf{F1}), and Semantic Similarity (\textbf{SSim}).
More details of the three metrics are shown in Appendix \ref{evaluation_metrics}.

\textbf{Implementation Details.}
We trained the Prompt-R1 agent using two large-scale LLMs as the environments: one with an overhead cost (GPT-4o-mini) and the other with zero cost (GPT-OSS-20B). More details are illustrated in Appendix \ref{implementation_metrics}.

\definecolor{oodcolor}{RGB}{127, 0, 255}      
\definecolor{avgcolor}{RGB}{0, 150, 255}      
\begin{table*}[t]
\vskip -0.15in
\begin{center}
\begin{small}
\fontsize{8pt}{8pt}\selectfont
\setlength{\tabcolsep}{1.686pt}
\setlength{\heavyrulewidth}{1.2pt}
\setlength{\lightrulewidth}{0.36pt}
\setlength{\arrayrulewidth}{0.36pt}
\begin{tabular}{l l c c c c c c c c c c}
\toprule
\textbf{Dataset} & \textbf{Metric} &
\multicolumn{2}{c}{\textbf{Baseline}} &
\multicolumn{1}{c}{\textbf{SFT}} &
\multicolumn{2}{c}{\textbf{CoT Reasoning}} &
\multicolumn{1}{c}{\textbf{GRPO}} &
\multicolumn{3}{c}{\textbf{APO (GPT-4o-mini)}} &
\multicolumn{1}{c}{\textbf{Ours}} \\
\cmidrule(lr){3-4}\cmidrule(lr){5-5}\cmidrule(lr){6-7}\cmidrule(lr){8-8}\cmidrule(lr){9-11}\cmidrule(lr){12-12}
& &
\makecell{Qwen3-4B} & \makecell{GPT-4o-mini} &
\makecell{Qwen3-4B} &
\makecell{Qwen3-4B} & \makecell{GPT-4o-mini} &
\makecell{Qwen3-4B} &
OPRO & TextGrad & GEPA & \makecell{\textbf{Prompt-R1} \textbf{($\Delta
\uparrow$)}}\\
\midrule
\multirow{2}{*}{\textbf{TriviaQA}} & \cellcolor{oodcolor!\coloropacity}\textbf{EM} & \cellcolor{oodcolor!\coloropacity}45.31 & \cellcolor{oodcolor!\coloropacity}63.28 & \cellcolor{oodcolor!\coloropacity}29.69 & \cellcolor{oodcolor!\coloropacity}46.88 & \cellcolor{oodcolor!\coloropacity}67.19 & \cellcolor{oodcolor!\coloropacity}51.56 & \cellcolor{oodcolor!\coloropacity}60.94 & \cellcolor{oodcolor!\coloropacity}54.69 & \cellcolor{oodcolor!\coloropacity}65.63 & \cellcolor{oodcolor!\coloropacity}\textbf{\increase{70.31}{7.03}} \\
 & \cellcolor{oodcolor!\coloropacity}\textbf{F1} & \cellcolor{oodcolor!\coloropacity}48.52 & \cellcolor{oodcolor!\coloropacity}71.34 & \cellcolor{oodcolor!\coloropacity}34.21 & \cellcolor{oodcolor!\coloropacity}50.26 & \cellcolor{oodcolor!\coloropacity}75.81 & \cellcolor{oodcolor!\coloropacity}55.98 & \cellcolor{oodcolor!\coloropacity}72.14 & \cellcolor{oodcolor!\coloropacity}67.59 & \cellcolor{oodcolor!\coloropacity}74.91 & \cellcolor{oodcolor!\coloropacity}\textbf{\increase{76.91}{5.57}} \\
\midrule
\multirow{2}{*}{\textbf{MathQA}} & \cellcolor{oodcolor!\coloropacity}\textbf{EM} & \cellcolor{oodcolor!\coloropacity}28.91 & \cellcolor{oodcolor!\coloropacity}46.09 & \cellcolor{oodcolor!\coloropacity}17.97 & \cellcolor{oodcolor!\coloropacity}27.34 & \cellcolor{oodcolor!\coloropacity}49.22 & \cellcolor{oodcolor!\coloropacity}46.88 & \cellcolor{oodcolor!\coloropacity}43.75 & \cellcolor{oodcolor!\coloropacity}44.53 & \cellcolor{oodcolor!\coloropacity}40.63 & \cellcolor{oodcolor!\coloropacity}\textbf{\increase{52.34}{6.25}} \\
 & \cellcolor{oodcolor!\coloropacity}\textbf{F1} & \cellcolor{oodcolor!\coloropacity}32.29 & \cellcolor{oodcolor!\coloropacity}54.04 & \cellcolor{oodcolor!\coloropacity}22.66 & \cellcolor{oodcolor!\coloropacity}30.60 & \cellcolor{oodcolor!\coloropacity}57.03 & \cellcolor{oodcolor!\coloropacity}54.43 & \cellcolor{oodcolor!\coloropacity}60.08 & \cellcolor{oodcolor!\coloropacity}61.46 & \cellcolor{oodcolor!\coloropacity}61.59 & \cellcolor{oodcolor!\coloropacity}\textbf{\increase{61.59}{7.55}} \\
\midrule
\multirow{2}{*}{\textbf{SQuAD v2}} & \cellcolor{oodcolor!\coloropacity}\textbf{EM} & \cellcolor{oodcolor!\coloropacity}6.25 & \cellcolor{oodcolor!\coloropacity}13.28 & \cellcolor{oodcolor!\coloropacity}5.47 & \cellcolor{oodcolor!\coloropacity}6.25 & \cellcolor{oodcolor!\coloropacity}14.06 & \cellcolor{oodcolor!\coloropacity}10.16 & \cellcolor{oodcolor!\coloropacity}10.94 & \cellcolor{oodcolor!\coloropacity}6.25 & \cellcolor{oodcolor!\coloropacity}13.28 & \cellcolor{oodcolor!\coloropacity}\textbf{\increase{19.53}{6.25}} \\
 & \cellcolor{oodcolor!\coloropacity}\textbf{F1} & \cellcolor{oodcolor!\coloropacity}16.09 & \cellcolor{oodcolor!\coloropacity}25.61 & \cellcolor{oodcolor!\coloropacity}16.18 & \cellcolor{oodcolor!\coloropacity}16.25 & \cellcolor{oodcolor!\coloropacity}25.73 & \cellcolor{oodcolor!\coloropacity}23.10 & \cellcolor{oodcolor!\coloropacity}26.67 & \cellcolor{oodcolor!\coloropacity}22.04 & \cellcolor{oodcolor!\coloropacity}25.52 & \cellcolor{oodcolor!\coloropacity}\textbf{\increase{29.28}{3.67}} \\
\midrule
\multirow{2}{*}{\textbf{XSum}} & \cellcolor{oodcolor!\coloropacity}\textbf{F1} & \cellcolor{oodcolor!\coloropacity}16.75 & \cellcolor{oodcolor!\coloropacity}24.35 & \cellcolor{oodcolor!\coloropacity}21.13 & \cellcolor{oodcolor!\coloropacity}8.33 & \cellcolor{oodcolor!\coloropacity}8.92 & \cellcolor{oodcolor!\coloropacity}21.98 & \cellcolor{oodcolor!\coloropacity}17.87 & \cellcolor{oodcolor!\coloropacity}24.88 & \cellcolor{oodcolor!\coloropacity}0.21 & \cellcolor{oodcolor!\coloropacity}\textbf{\increase{25.76}{1.41}} \\
 & \cellcolor{oodcolor!\coloropacity}\textbf{SSim} & \cellcolor{oodcolor!\coloropacity}31.73 & \cellcolor{oodcolor!\coloropacity}60.56 & \cellcolor{oodcolor!\coloropacity}55.65 & \cellcolor{oodcolor!\coloropacity}6.81 & \cellcolor{oodcolor!\coloropacity}53.94 & \cellcolor{oodcolor!\coloropacity}52.09 & \cellcolor{oodcolor!\coloropacity}29.47 & \cellcolor{oodcolor!\coloropacity}28.45 & \cellcolor{oodcolor!\coloropacity}0.37 & \cellcolor{oodcolor!\coloropacity}\textbf{\increase{63.02}{2.46}} \\
\midrule
\multirow{3}{*}{\textbf{Average}} & \cellcolor{avgcolor!\coloropacity}\textbf{EM}   & \cellcolor{avgcolor!\coloropacity}26.82 & \cellcolor{avgcolor!\coloropacity}40.88 & \cellcolor{avgcolor!\coloropacity}17.71 & \cellcolor{avgcolor!\coloropacity}26.82 & \cellcolor{avgcolor!\coloropacity}43.49 & \cellcolor{avgcolor!\coloropacity}36.20 & \cellcolor{avgcolor!\coloropacity}38.54 & \cellcolor{avgcolor!\coloropacity}35.16 & \cellcolor{avgcolor!\coloropacity}39.85 & \cellcolor{avgcolor!\coloropacity}\textbf{\increase{47.39}{6.51}} \\
 & \cellcolor{avgcolor!\coloropacity}\textbf{F1}   & \cellcolor{avgcolor!\coloropacity}28.41 & \cellcolor{avgcolor!\coloropacity}43.84 & \cellcolor{avgcolor!\coloropacity}23.55 & \cellcolor{avgcolor!\coloropacity}26.36 & \cellcolor{avgcolor!\coloropacity}41.87 & \cellcolor{avgcolor!\coloropacity}38.87 & \cellcolor{avgcolor!\coloropacity}44.19 & \cellcolor{avgcolor!\coloropacity}43.99 & \cellcolor{avgcolor!\coloropacity}40.56 & \cellcolor{avgcolor!\coloropacity}\textbf{\increase{48.39}{4.55}} \\
 & \cellcolor{avgcolor!\coloropacity}\textbf{SSim} & \cellcolor{avgcolor!\coloropacity}31.73 & \cellcolor{avgcolor!\coloropacity}60.56 & \cellcolor{avgcolor!\coloropacity}55.65 & \cellcolor{avgcolor!\coloropacity}6.81  & \cellcolor{avgcolor!\coloropacity}53.94 & \cellcolor{avgcolor!\coloropacity}52.09 & \cellcolor{avgcolor!\coloropacity}29.47 & \cellcolor{avgcolor!\coloropacity}28.45 & \cellcolor{avgcolor!\coloropacity}0.37  & \cellcolor{avgcolor!\coloropacity}\textbf{\increase{63.02}{2.46}} \\
\bottomrule
\end{tabular}
\end{small}
\end{center}
\vskip -0.1in
\caption{Comparison of the baselines and Prompt-R1 on four OOD datasets, including \textbf{TriviaQA} (\textbf{multi-hop reasoning}), \textbf{MathQA} (\textbf{mathematical computation}), \textbf{SQuAD v2} (\textbf{standard QA}), and \textbf{XSum} (\textbf{text generation}).}
\label{tab:prompt-r1-ood}
\end{table*}

\subsection{Main Results (RQ1)}
\label{4-1}
As illustrated in Table \ref{tab:prompt-r1-train}, our proposed Prompt-R1 significantly improves the performance of the baseline (GPT-4o-mini) across eight datasets and outperforms other baselines, including SFT, CoT, GRPO, and APO methods. It excels in the four tasks, including multi-hop reasoning, mathematical reasoning, QA, and text generation. It achieves the largest gains in multi-hop reasoning, enhances stability in mathematical reasoning, and performs exceptionally well in knowledge retrieval tasks such as PopQA. In text generation, it consistently improves quality, demonstrating strong robustness. Overall, these results demonstrate that Prompt-R1 has three major advantages: \textit{(i)} consistent improvement across tasks, confirming its broad applicability; \textit{(ii)} significantly enhances the performance of baseline LLMs in complex reasoning tasks; and \textit{(iii)} superior stability compared to other 
baseline methods, preventing performance collapse. 

\begin{figure}[t]
\centering
\includegraphics[width=0.486\textwidth]{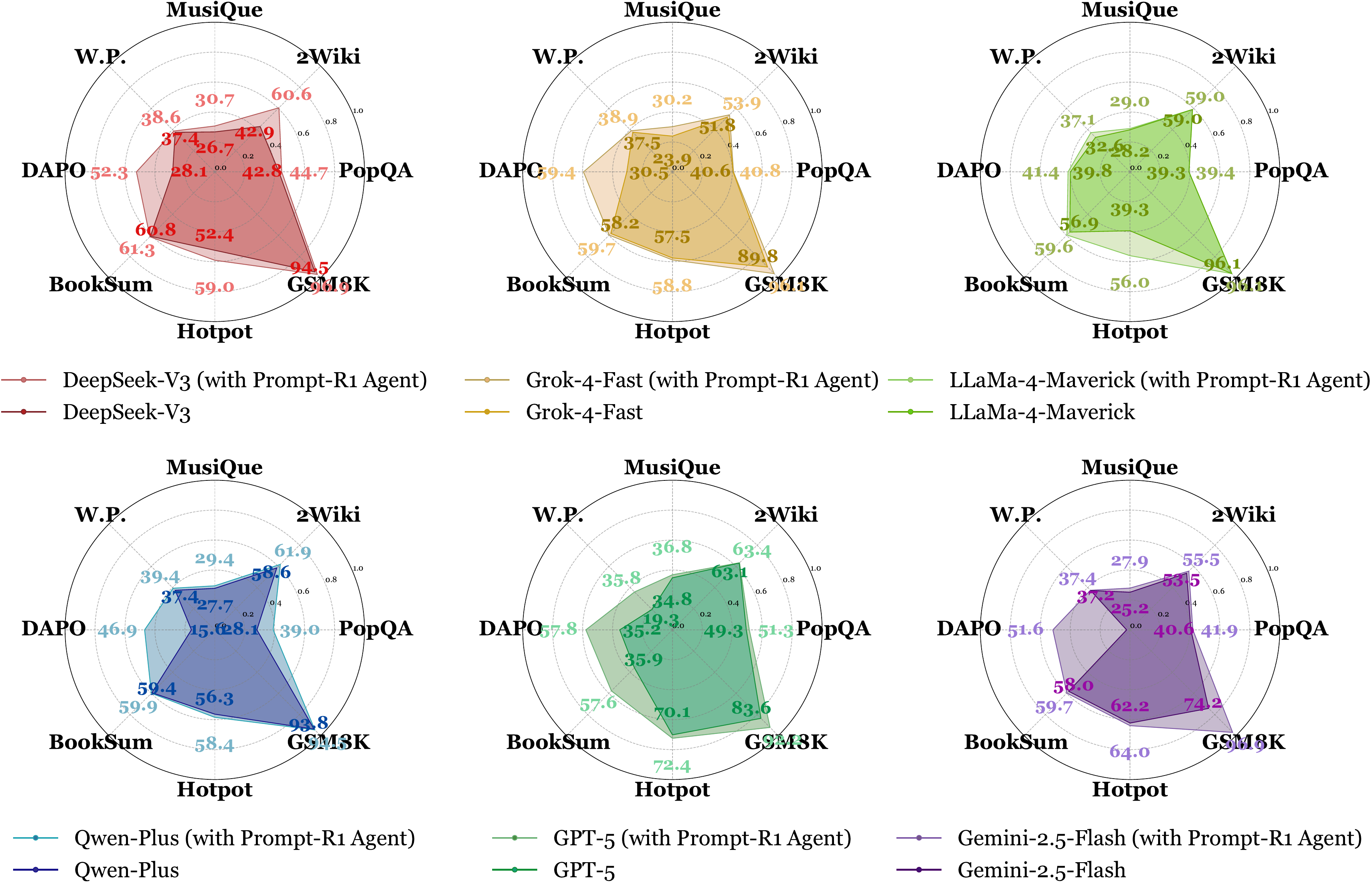} 
\caption{The performance of six LLMs without Prompt-R1 agent is compared to that with Prompt-R1 on eight datasets, using appropriate metrics. F1: multi-hop reasoning (2Wiki, Hotpot) and standard QA (MusiQue, PopQA); EM: mathematical computation (GSM8K, DAPO); and SSim: text generation (BookSum, W.P.).}
\label{fig:6-models-8-datasets-radar}
\end{figure}

\subsection{Generalization Results (RQ2)}
\label{4-2}
As shown in Table \ref{tab:prompt-r1-ood}, Prompt-R1 consistently outperforms the baseline methods across the four public out-of-distribution datasets, with significant improvements in multi-hop reasoning and mathematical computation tasks, particularly in EM and F1 metrics. While standard QA tasks show some improvement, they are limited by baseline performance, and text generation shows moderate gains in structural similarity. Among the baselines, large-scale LLMs exhibit stronger zero-shot reasoning, while supervised fine-tuning faces overfitting and distribution shift issues. CoT reasoning shows inconsistent results, and automatic prompting optimization methods vary across tasks, reflecting the complexity of the strategy. Overall, Prompt-R1 demonstrates task-independent effectiveness, especially in reasoning-intensive tasks, and offers competitive performance without the need for LLM fine-tuning or a large-scale annotated dataset.

\begin{figure}[t]
\centering
\includegraphics[width=0.476\textwidth]{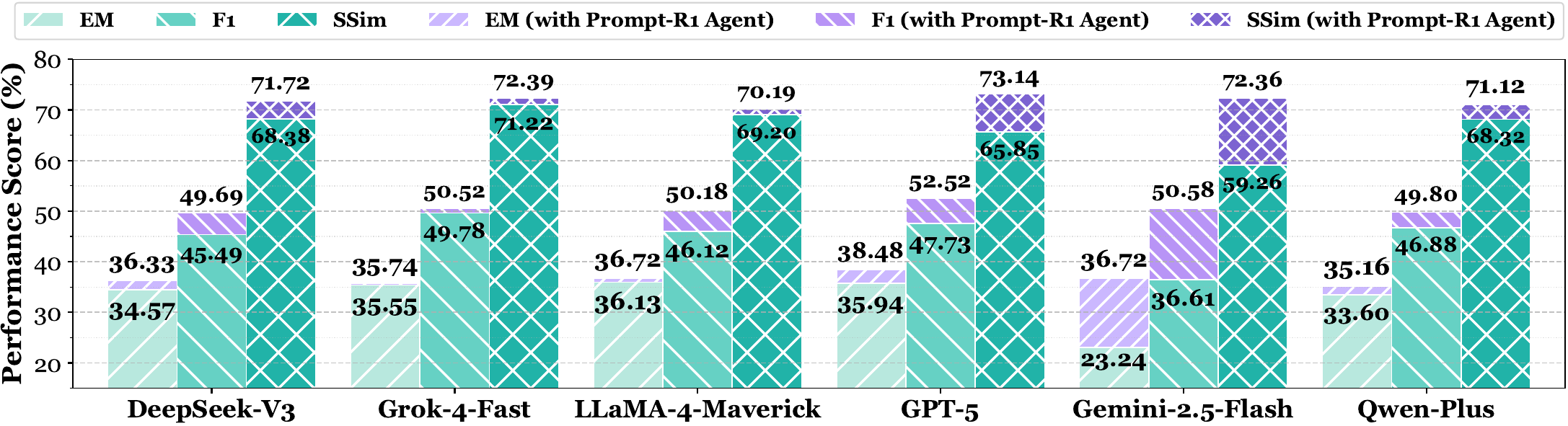} 
\caption{Comparison of the average values of the three EM, F1, and SSim scores for the selected six LLM baselines on OOD datasets, as well as the average of these scores for these LLMs using the Prompt-R1 agent.}
\label{fig:6-models-OOD}
\end{figure}

\subsection{Work with Other LLMs (RQ3)}
\label{4-3}
\label{exp-proof3}
The comparisons of six LLMs (e.g., Deepseek-V3, Grok-4-fast, LLaMA-4-Maverick, GPT-5, Gemini-2.5-flash, and Qwen-Plus) and their collaboration with the Prompt-R1 agent across 8 in-distribution and 4 OOD datasets are shown in Figures \ref{fig:6-models-8-datasets-radar} and \ref{fig:6-models-OOD}. As shown in Figure \ref{fig:6-models-8-datasets-radar}, on in-distribution datasets, the Prompt-R1 agent boosts performance in multi-hop reasoning tasks, with limited gains in mathematical reasoning, reflecting the baseline LLM’s strengths. The agent’s impact varies across LLMs, with greater improvements in low-baseline tasks and diminishing returns in high-baseline tasks. As shown in Figure \ref{fig:6-models-OOD}, on OOD datasets, incorporating the Prompt-R1 agent leads to substantial improvements in EM, F1, and SSim, especially in SSim, enhancing semantic coherence. LLMs with weaker baselines benefit most, confirming the Prompt-R1 agent’s ability to address reasoning weaknesses and maintain consistent performance across LLMs.

\begin{table*}[t]
\vskip -0.15in
\begin{center}
\fontsize{8pt}{8pt}\selectfont
\setlength{\tabcolsep}{6.86pt}
\renewcommand{\arraystretch}{1.0}
\setlength{\heavyrulewidth}{1.2pt}
\setlength{\lightrulewidth}{0.36pt}
\setlength{\arrayrulewidth}{0.36pt}
\begin{tabular}{l c c c c c c c c c c c c c}
\toprule
\textbf{Dataset} & \multicolumn{2}{c}{\textbf{2Wiki}} & \multicolumn{2}{c}{\textbf{Hotpot}} & \multicolumn{2}{c}{\textbf{GSM8K}} & \multicolumn{2}{c}{\textbf{DAPO}} & \multicolumn{2}{c}{\textbf{MusiQue}} & \multicolumn{2}{c}{\textbf{PopQA}} \\
\cmidrule(lr){2-3}\cmidrule(lr){4-5}\cmidrule(lr){6-7}\cmidrule(lr){8-9}\cmidrule(lr){10-11}\cmidrule(lr){12-13}
& \textbf{EM} & \textbf{F1} & \textbf{EM} & \textbf{F1} & \textbf{EM} & \textbf{F1} & \textbf{EM} & \textbf{F1} & \textbf{EM} & \textbf{F1} & \textbf{EM} & \textbf{F1} \\
\midrule
w/o Env. & 41.41 & 43.48 & 21.88 & 25.96 & 95.31 & 95.31 & 25.00 & 25.00 & 2.34 & 11.63 & 8.59 & 10.85 \\
w/o R.L. & 1.56 & 8.86 & 20.31 & 27.42 & 84.38 & 89.41 & 21.09 & 22.22 & 5.47 & 10.90 & 3.91 & 9.28 \\
w/o Agent & 33.59 & 36.57 & 33.59 & 40.07 & 83.59 & 86.72 & 18.75 & 18.76 & 14.06 & 22.06 & 25.78 & 30.75 \\
\textbf{Prompt-R1 (Full)} & \textbf{48.44} & \textbf{54.41} & \textbf{44.53} & \textbf{52.31} & \textbf{97.66} & \textbf{97.66} & \textbf{26.56} & \textbf{26.56} & \textbf{18.75} & \textbf{26.31} & \textbf{28.13} & \textbf{33.77} \\
\midrule
\textbf{Dataset} & \multicolumn{2}{c}{\textbf{BookSum}} & \multicolumn{2}{c}{\textbf{W.P.}} & \multicolumn{2}{c}{\textbf{MathQA}} & \multicolumn{2}{c}{\textbf{SQuAD v2}} & \multicolumn{2}{c}{\textbf{TriviaQA}} & \multicolumn{2}{c}{\textbf{XSum}} \\
\cmidrule(lr){2-3}\cmidrule(lr){4-5}\cmidrule(lr){6-7}\cmidrule(lr){8-9}\cmidrule(lr){10-11}\cmidrule(lr){12-13}
& \textbf{F1} & \textbf{SSim} & \textbf{F1} & \textbf{SSim} & \textbf{EM} & \textbf{F1} & \textbf{EM} & \textbf{F1} & \textbf{EM} & \textbf{F1} & \textbf{F1} & \textbf{SSim} \\
\midrule
w/o Env. & 23.34 & 56.84 & 18.67 & 37.83 & 49.22 & 57.94 & 10.16 & 18.12 & 39.84 & 43.54 & 21.41 & 59.87 \\
w/o R.L. & 19.38 & 58.80 & 17.11 & 36.65 & 48.44 & 58.07 & 5.47 & 12.89 & 49.22 & 56.23 & 23.61 & 61.83 \\
w/o Agent & 20.45 & 57.01 & 19.48 & 35.31 & 46.09 & 54.04 & 13.28 & 25.61 & 63.28 & 71.34 & 24.35 & 60.56 \\
\textbf{Prompt-R1 (Full)} & \textbf{26.50} & \textbf{60.87} & \textbf{22.11} & \textbf{38.54} & \textbf{52.34} & \textbf{61.59} & \textbf{19.53} & \textbf{29.28} & \textbf{70.31} & \textbf{76.91} & \textbf{25.76} & \textbf{63.02} \\
\bottomrule
\end{tabular}
\end{center}
\vskip -0.1in
\caption{Ablation study of Prompt-R1 (GPT-4o-mini as environment) on twelve datasets, including Prompt-R1 (Full), without LLM-as-Environment (w/o Env.), without RL (w/o R.L.), and without Prompt-R1 agent (w/o Agent).}
\label{tab:ablation-extended}
\end{table*}

\begin{figure*}[t]
    \centering
    \includegraphics[width=0.2\textwidth]{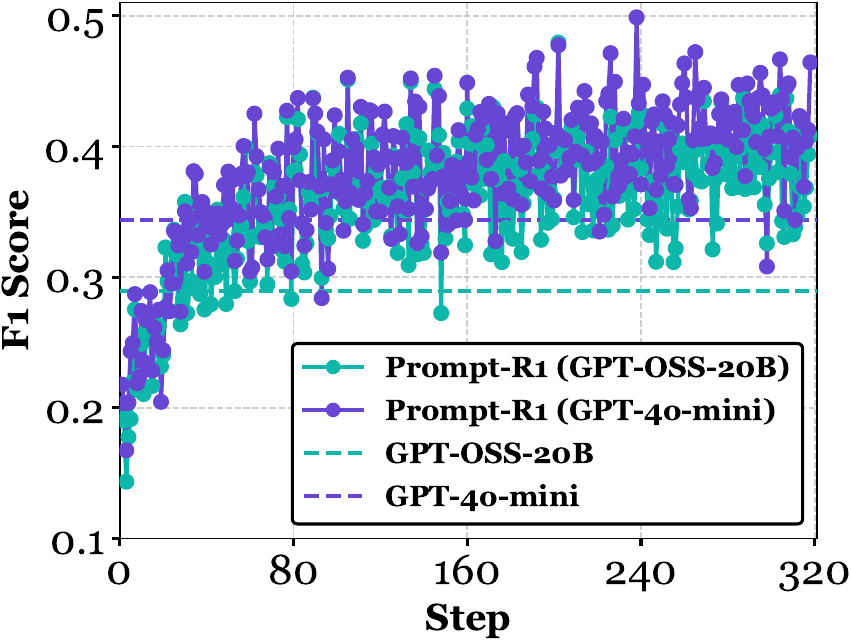}\hfill
    \includegraphics[width=0.2\textwidth]{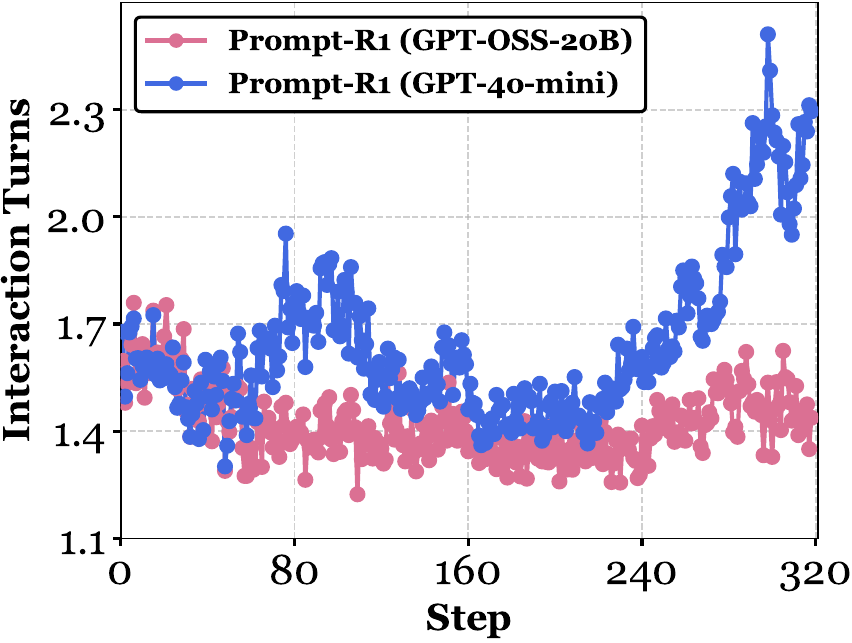}\hfill
    \includegraphics[width=0.19\textwidth]{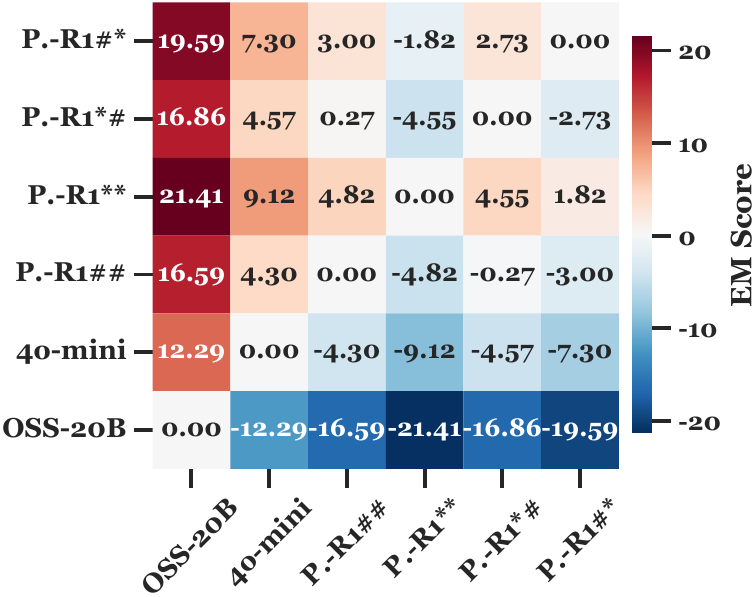}\hfill
    \includegraphics[width=0.189\textwidth]{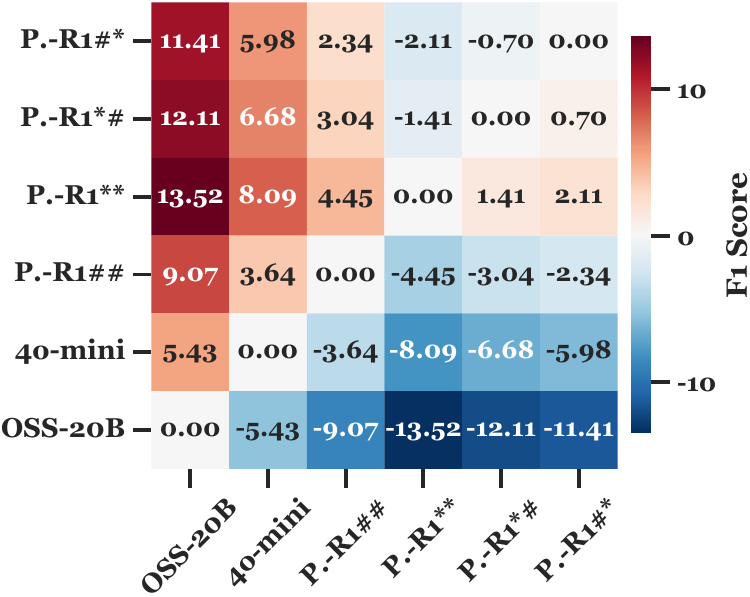}\hfill
    \includegraphics[width=0.185\textwidth]{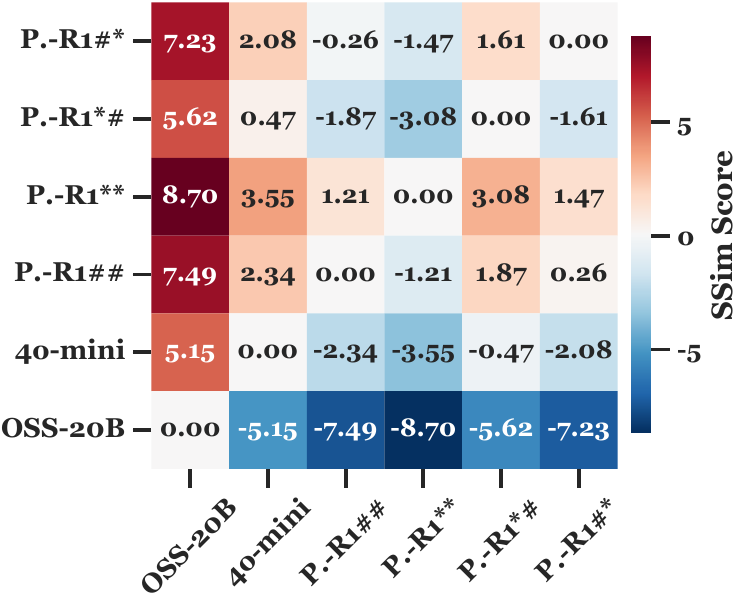}
    \parbox{0.2\textwidth}{\centering\small (a) Training process}\hfill
    \parbox{0.2\textwidth}{\centering\small (b) Interaction Turns}\hfill
    \parbox{0.2\textwidth}{\centering\small (c) Average EM}\hfill
    \parbox{0.2\textwidth}{\centering\small (d) Average F1}\hfill
    \parbox{0.2\textwidth}{\centering\small (e) Average SSim} 
    \vspace{-5.6mm}
    \caption{(a-b) Training process and interaction turns of Prompt-R1 agent with different environments. (c-e) Comparison of different Prompt-R1 agents with environments on average EM, average F1, and average SSim.}
    \label{fig: cost}
\end{figure*}

\begin{table}[t]
\begin{center}
\fontsize{8pt}{8pt}\selectfont
\setlength{\tabcolsep}{1.66pt}
\renewcommand{\arraystretch}{1.0}
\setlength{\heavyrulewidth}{1.2pt}
\setlength{\lightrulewidth}{0.36pt}
\setlength{\arrayrulewidth}{0.36pt}
\begin{tabular}{l c c c c c c c c}
\toprule
\textbf{Dataset} & \multicolumn{2}{c}{\textbf{MathQA}} & \multicolumn{2}{c}{\textbf{SQuAD v2}} & \multicolumn{2}{c}{\textbf{TriviaQA}} & \multicolumn{2}{c}{\textbf{XSum}} \\
\cmidrule(lr){2-3}\cmidrule(lr){4-5}\cmidrule(lr){6-7}\cmidrule(lr){8-9}
& \textbf{EM} & \textbf{F1} & \textbf{EM} & \textbf{F1} & \textbf{EM} & \textbf{F1} & \textbf{F1} & \textbf{SSim} \\
\midrule
OSS & \underline{51.56} & \underline{59.48} & 7.81 & 18.75 & 50.78 & 59.37 & 22.77 & 59.48 \\
4o-mini & 46.09 & 54.04 & \underline{13.28} & \underline{25.61} & \underline{63.28} & \underline{71.34} & \underline{24.35} & \underline{60.56} \\
\midrule
Prompt-R1\textsuperscript{\#\#} & 51.56 & 61.07 & 10.16 & 21.28 & 50.78 & 59.60 & 24.14 & 61.83 \\
Prompt-R1\textsuperscript{**} & 52.34 & 61.59 & \textbf{19.53} & \textbf{29.28} & \textbf{70.31} & \textbf{76.91} & \textbf{25.76} & \textbf{63.02} \\
Prompt-R1\textsuperscript{*\#} & \textbf{53.91} & \textbf{62.63} & 8.59 & 20.13 & 51.56 & 60.05 & 22.95 & 60.84 \\
Prompt-R1\textsuperscript{\#*} & 52.34 & 60.29 & 17.97 & 27.96 & 67.19 & 73.11 & 25.82 & 61.83 \\
\bottomrule
\end{tabular}
\end{center}
\caption{Comparison of different Prompt-R1 agents with environments and two baseline LLMs (GPT-OSS-20B: OSS and GPT-4o-mini: 4o-mini) on four OOD datasets. Prompt-R1\textsuperscript{**} (P.-R1\textsuperscript{**}) means both agent and environment are GPT-4o-mini; Prompt-R1\textsuperscript{\#\#} (P.-R1\textsuperscript{\#\#}) means both agent and environment are GPT-OSS-20B; Prompt-R1\textsuperscript{*\#} (P.-R1\textsuperscript{*\#}) means the agent is GPT-4o-mini trained and the environment is GPT-OSS-20B; Prompt-R1\textsuperscript{\#*} (P.-R1\textsuperscript{\#*}) means the agent is GPT-OSS-20B trained and the environment is GPT-4o-mini. Underlined values indicate better performance between the two baselines.}
\label{tab:agent-env-comparison}
\vskip -0.1in
\end{table}

\subsection{Ablation Study (RQ4)}
\label{4-4}
As shown in Table \ref{tab:ablation-extended}, Prompt-R1 (Full) performs optimally across all datasets, highlighting the strong synergy of environment, reinforcement learning, and the agent. Removing any single component significantly reduces performance, with reinforcement learning most affected by complex reasoning tasks. The environment has notably less impact on mathematical computation, showing clear task-dependent dependencies. In addition, Prompt-R1 (Full) exhibits more stable and consistent performance on the four OOD datasets, reflecting strong cross-task generalization ability. Ablation experiments show that the performance effectiveness of Prompt-R1 stems from the close coupling of environment interaction, the learning mechanisms, and agent decision-making, with varying component dependencies across tasks, and the agent-environment configuration should be carefully selected based on the task and requirement.


\subsection{Different Environment Training (RQ5)}
\label{4-5}
We trained the Prompt-R1 agent in two environments: GPT-4o-mini (via official API) and GPT-OSS-20B (locally deployed). Results in Table \ref{tab:agent-env-comparison} and Figure \ref{fig: cost} reveal five insights:
\textit{(i)} The zero-cost LLM (GPT-OSS-20B) as training environment achieves performance close to the overhead-cost LLM (GPT-4o-mini) alternative;
\textit{(ii)} With GPT-4o-mini, Prompt-R1 improves faster but is less stable, while GPT-OSS-20B shows steadier convergence and more efficient feedback;
\textit{(iii)} The agent-environment interaction framework significantly enhances reasoning, especially for weaker LLMs;
\textit{(iv)} The environment, as both reasoning generator and feedback verifier, has a decisive impact on performance;
\textit{(v)} Collaboration with the training environment enhances performance, indicating that RL can strengthen agent and LLMs' cooperation.

\section{Conclusion}
In this work, we propose Prompt-R1 to replace humans in interacting more effectively with LLMs. We find 3 key similarities between Prompt-R1 agent and humans: \textit{(i)} learning from others' knowledge through communication, \textit{(ii)} improving communication efficiency with continuous interactions, and \textit{(iii)} communication skills can be transferred to engage effectively with others. We hope Prompt-R1 offers a new approach to human-LLM interaction.
\newpage

\section*{Limitations}
In Appendix \ref{Limitations}, we discuss the limitations of the proposed Prompt-R1. Furthermore, we also provide the case study in Appendix \ref{case_study}.

\section*{Ethical Considerations}
This research utilizes publicly available datasets without involving sensitive or personally identifiable data. We believe it does not violate any ethics.

\section*{Acknowledgments}
This work was supported by the National Key Research and Development Program of China [grant number 2023YFF0905404].




\bibliography{custom}

\begin{figure*}[h]
\centering
\includegraphics[width=0.986\textwidth]{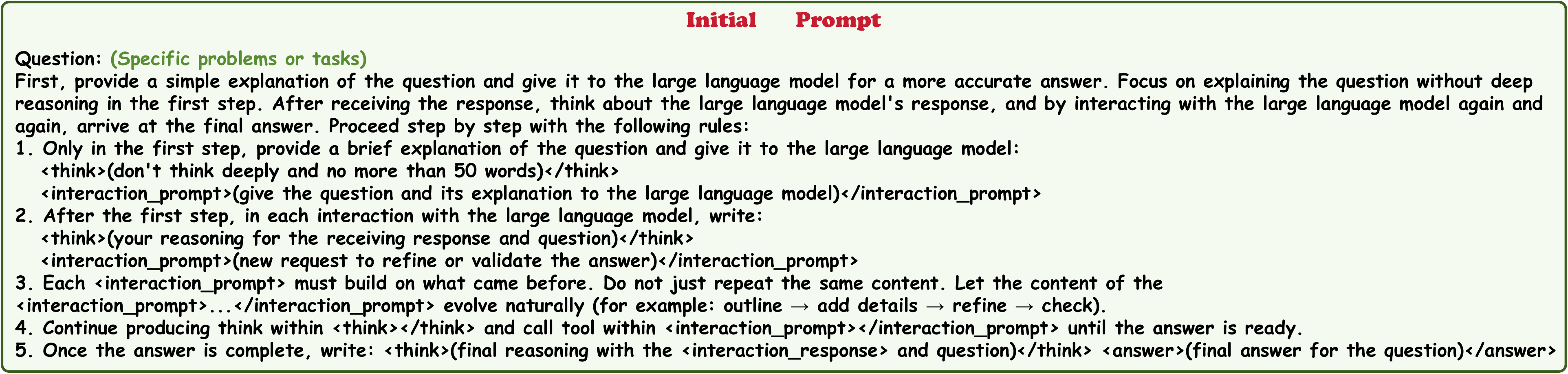}
\caption{The initial prompt template is utilized for the agent (small-scale LLM) to communicate with the environment (Large-scale LLM).}
\label{fig: Initial_Prompt}
\end{figure*}

\begin{figure*}[h!]
\centering
\includegraphics[width=0.986\textwidth]{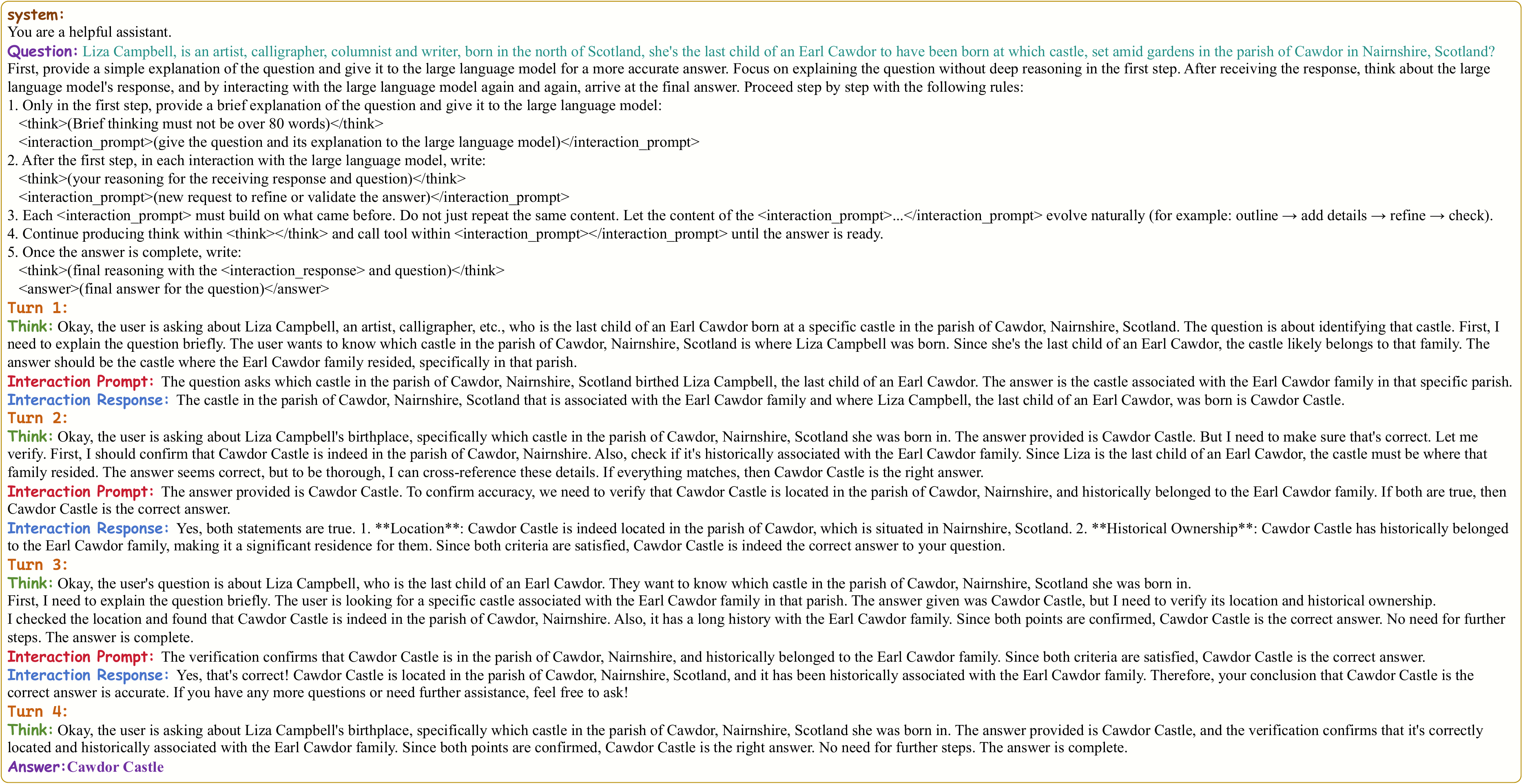}
\caption{An illustration of the multi-turn interactions of agent (small-scale LLM) and environment (large-scale LLM), showing how the agent alternates between reflection and communication, through verification and targeted evaluation, progressively refining the reasoning process until the answer is obtained.}
\label{fig: example}
\end{figure*}

\newpage
\appendix

\section*{Appendix}
\label{sec:appendix}

\section{Prompts Used in Prompt-R1}
\subsection{Initial Prompt}
As shown in Figure \ref{fig: Initial_Prompt}, we adopt the initial prompt template for the agent (small-scale LLM), where the agent alternates between internal reasoning in \texttt{<think>} and communicates with the large-scale LLM by \texttt{<interaction\_prompt>}. The prompts evolve step by step until the solution is finalized with a concluding \texttt{<think>} and \texttt{<answer>}.

\subsection{Multi-turn Prompt Interaction Process}
A complete multi-turn interaction between the agent (small-scale LLM) and the environment (large-scale LLM) is illustrated in Figure~\ref{fig: example}. The process begins with an initial interpretation of the user’s question, which establishes the context for subsequent reasoning and guides the trajectory of analysis. The agent then engages in iterative exchanges: after each response, it reflects on the output, generates prompts for answering questions, and verifies essential aspects of the information received. Through successive rounds of clarification, the reasoning gradually converges, and the agent produces a coherent, grounded final answer.

\section{Theoretical Proof}
\subsection{Proof of Proposition 1}
\label{proof1}
\textbf{Proposition 1.} \textit{Multi-turn interactions of the small-scale LLM and large-scale LLM can better solve problems.}
\begin{proof}
Let the answer set be $\mathcal{A}=\{1,\ldots,M\}$, and assume the true label $Y\in\mathcal{A}$.  
At the interaction round $t$, the agent $S$ (small-scale LLM) chooses an action $a_t$ based on the previous prompt interaction history with the environment $L$:
\begin{equation}
H_{t-1}=\{(a_1,o_1),\ldots,(a_{t-1},o_{t-1})\},
\end{equation}
where $H_{t-1}$ is the interaction history up to round $t-1$,  
$a_s$ is the prompt at round $s$, and $o_s$ is the corresponding observation.

The environment (large-scale LLM) $L$ produces an observation conditioned on $Y$ as:
\begin{equation}
o_t\sim K_{a_t}(\cdot\mid Y),
\end{equation}
where $K_{a_t}(\cdot\mid Y)$ is the class-conditional observation law indexed by action $a_t$. The history is then updated:
\begin{equation}
H_t = H_{t-1}\oplus(a_t,o_t),
\end{equation}
where $\oplus$ is appending the ordered pair $(a_t,o_t)$ to the history. Define the posterior vector and the Bayes accuracy function as:
\begin{equation}
\boldsymbol{\pi}_t(y)\triangleq \mathbb{P}(Y=y\mid H_t),
A(H_t)\triangleq \max_{y\in\mathcal{A}}\boldsymbol{\pi}_t(y),
\end{equation}
where $\boldsymbol{\pi}_t(y)$ is the posterior probability of class $y$ given $H_t$,  
and $A(H_t)$ is the Bayes accuracy under $0$-$1$ loss.
Introduce the Bayes risk potential function to measure uncertainty, which is calculated as:
\begin{equation}
V(H_t)\triangleq 1-A(H_t)=1-\max_{y\in\mathcal{A}}\boldsymbol{\pi}_t(y),
\end{equation}
where $V(H_t)$ is the Bayes risk potential, smaller values indicate lower uncertainty.

\textit{(i) Posterior martingale and expectation contraction.}  
By Bayes’ rule, the posterior vector satisfies the martingale property:
\begin{equation}
\mathbb{E}\!\left[\boldsymbol{\pi}_t(y)\mid \mathscr{F}_{t-1}\right]=\boldsymbol{\pi}_{t-1}(y),
\end{equation}
where $\mathscr{F}_{t-1}=\sigma(H_{t-1})$ is the natural filtration generated by the prompt interaction history.
Consider the concave potential function over the probability simplex, which is calculated as follows:
\begin{equation}
\phi(\boldsymbol{p})\triangleq 1-\max_{y}p_y,\qquad \boldsymbol{p}\in\Delta^{M-1},
\end{equation}
where $\boldsymbol{p}$ is a probability vector on $\mathcal{A}$,  
and $\Delta^{M-1}$ is the $(M-1)$-simplex of such vectors.
Applying Jensen’s inequality gives the contraction of expected risk:
\begin{equation}
\mathbb{E}[V(H_t)\mid \mathscr{F}_{t-1}] \le V(H_{t-1}),
\end{equation}
where the inequality is strict whenever $K_{a_t}(\cdot\mid Y)$ is information-bearing.

\textit{(ii) Monotone improvement over multiple turns.}  
Taking the unconditional expectation and iterating the relation yields:
\begin{equation}
\mathbb{E}[V(H_t)] \le \mathbb{E}[V(H_{t-1})] \le \cdots \le \mathbb{E}[V(H_0)].
\end{equation}

Define the accuracy gain at round $t$ by:
\begin{equation}
\Delta_t \triangleq \mathbb{E}[V(H_{t-1})-\mathbb{E}[V(H_t)\mid \mathscr{F}_{t-1}]] \ge 0,
\end{equation}
where $\Delta_t$ is the expected one-step reduction of the Bayes risk potential.
Hence, the expectation of Bayes risk after $t$ rounds satisfies:
\begin{equation}
\mathbb{E}[V(H_t)] = \mathbb{E}[V(H_0)] - \sum_{s=1}^{t}\Delta_s,
\end{equation}
where each $\Delta_s$ accumulates the expected risk decrease at round $s$.
Substituting into the definition of accuracy, we obtain
\begin{equation}
\mathbb{E}[A(H_t)] = 1-\mathbb{E}[V(H_0)] + \sum_{s=1}^{t}\Delta_s,
\end{equation}
where $\mathbb{E}[A(H_t)]$ is the expected Bayes accuracy after $t$ rounds.

\textit{(iii) Asymptotic correctness under identifiability.}  
Suppose the family $\{K_{a_t}(\cdot\mid y):y\in\mathcal{A}\}$ is identifiable infinitely often.  
Then, by Doob’s martingale convergence theorem, we obtain:
\begin{equation}
V(H_t) \xrightarrow[t\to\infty]{\text{a.s.}} 0,\qquad
A(H_t) \xrightarrow[t\to\infty]{\text{a.s.}} 1,
\end{equation}
where $a.s.$ abbreviates almost sure convergence with respect to the data law.
Taking the expectation over the convergence yields:
\begin{equation}
\lim_{t\to\infty}\mathbb{E}[A(H_t)] = 1
\end{equation}
where the limit is taken under the assumed identifiability condition.

In conclusion, multi-turn small-large LLM interactions strictly decrease the Bayes risk whenever observations are informative, and consequently increase the expected accuracy. With repeated informative interactions, the expected accuracy approaches one. Therefore, multi-turn interaction ensures higher accuracy and superiority in solving problems.  
\end{proof}

\subsection{Proof of Proposition 2}
\label{proof2}
\textbf{Proposition 2.} \textit{Reinforcement learning can make small-scale LLMs better guide large-scale LLMs to complete tasks.}
\begin{proof} 
Let the question \( q \) be the input, and the true answer \( Y \) be a random variable, \( Y \in \mathcal{A} \), with the policy of the small-scale LLM \( \pi_{\tau} \) and the prompt sequence generated by the small-scale LLM \( \tau = (a_1, a_2, \dots, a_T) \). The joint distribution between the small-scale LLM-generated prompt sequence and the large-scale LLM output \( Y \) is represented as:
\begin{equation}
P(\tau, Y | q) = \pi_{\tau}(\tau | q) P_L(Y | \tau, q)
\end{equation}
where \( \pi_{\tau}(\tau | q) \) represents the probability of generating the prompt sequence \( \tau \) given the task \( q \), and \( P_L(Y | \tau, q) \) is the conditional distribution of the large-scale LLM generating the answer \( Y \) given the prompt sequence \( \tau \). 
To optimize the policy of the small-scale LLM, we use reinforcement learning to maximize the expected reward and minimize the KL divergence for regularization, with the optimization objective:
\begin{equation}
J(\pi_{\tau}) = \mathbb{E}_{q} [ R ] - \beta D_{\mathrm{KL}}( \pi_{\tau} \| \pi_{\mathrm{ref}} )
\end{equation}
where \( \mathbb{E}_{q} [ R ] \) represents the expected reward based on the reward signal, \( D_{\mathrm{KL}} \) is the KL divergence used for regularizing the policy update, and \( \beta \) is the regularization coefficient controlling the impact of the KL divergence. By maximizing the expected reward and minimizing the KL divergence, reinforcement learning optimization allows the small-scale LLM’s policy to generate effective prompt sequences, thereby improving task accuracy.

To prove that reinforcement learning optimization can improve the accuracy of the small-scale LLM, we utilize negative log-likelihood to measure the accuracy of the current policy. Let \( U(\pi_{\tau}) \) be the energy term under the current policy:
\begin{equation}
U(\pi_{\tau}) = \mathbb{E}_{q,Y} \left[ - \log P_L(Y | q, \tau) \right]. 
\end{equation}
According to the policy gradient theorem, we know that maximizing the expected reward and minimizing the KL divergence through RL can optimize the policy, thereby improving accuracy. Using the policy gradient theorem, we have:
\begin{equation}
\nabla_{\pi} J(\pi) = \mathbb{E}_{q} \left[ \nabla_{\pi} \log \pi_{\tau} \cdot R \right] - \beta \mathbb{E}_{q} \left[ \nabla_{\pi} D_{\mathrm{KL}} \right].
\end{equation}
This provides the mathematical framework for optimizing the small-scale LLM's policy. By adjusting the policy \( \pi_{\tau} \), the small-scale LLM can better guide the large-scale LLM to generate accurate results. By combining the reinforcement learning optimization objective and the policy gradient theorem, we can derive the relationship between the accuracy of the small-scale LLM's policy and the energy term. Using the log-sum inequality, we have:

\begin{equation}
\text{Acc}(\pi_{\tau}) \geq \exp(-U(\pi_{\tau})).
\end{equation}

This shows that through reinforcement learning optimization, the optimal policy will maximize accuracy and minimize the energy term \( U(\pi_{\tau}) \), thereby improving task accuracy.
Through reinforcement learning optimization, we obtain an optimal policy \( \pi^* \) that significantly improves the task accuracy of the small-scale LLM-generated prompt sequence:
\begin{equation}
\text{Acc}(\pi^*) \geq \text{Acc}(\pi_{\tau}).
\end{equation}
This proves that the optimal policy obtained through reinforcement learning optimization can significantly improve task accuracy, thereby helping the small-scale LLM better guide the large-scale LLM to complete the task.
\end{proof}

\subsection{Proof of Proposition 3}
\label{proof3}
\textbf{Proposition 3.} \textit{The agent can enhance not only the LLM used for its training but also other LLMs.}
\begin{proof} 
Let the answer set be \(\mathcal{A}=\{1,\ldots,M\}\), and the true label be \(Y\in\mathcal{A}\).  
At interaction round \(t\), the agent (small-scale LLM) chooses an action (prompt) \(a_t\); the environment (large-scale LLM)  \(L\) produces an observation conditioned on \(Y\) as:
\begin{equation}
o_t \sim K^{L}_{a_t}(\cdot\mid Y),
\end{equation}
where \(K^{L}_{a}(\cdot\mid Y)\) is the class-conditional observation kernel indexed by action \(a\) under environment \(L\).  
The history then updates as
\begin{equation}
H_t=H_{t-1}\oplus(a_t,o_t),\qquad H_0=\varnothing,
\end{equation}
where \(\oplus\) denotes appending the ordered pair \((a_t,o_t)\) to the multi-turn prompt interaction history.  
In another environment \(L'\), the observation kernel is denoted by \(K^{L'}_{a}(\cdot\mid Y)\).  
The policy of the agent is denoted by \(\pi\), which selects \(a_t\) at round \(t\) based on the past history \(H_{t-1}\).  
A trajectory of length \(T\) is denoted by \(\tau=(a_1,o_1,\ldots,a_T,o_T)\).  
Given task \(q\) and ground-truth \(Y\), the joint distribution under environment \(E\in\{L,L'\}\) with policy \(\pi\) is
\begin{equation}
\begin{split}
P^{\pi}_{E}(\tau,Y\mid q) 
&= \big( \prod_{t=1}^{T} \pi(a_t\mid H_{t-1},q) \big) \\
&\hspace{-2em} \big( \prod_{t=1}^{T} K^{E}_{a_t}(o_t\mid Y) \big) \times P(Y\mid q).
\end{split}
\end{equation}
The utility function of a trajectory is denoted by \(U(\tau,Y)\in[0,1]\) (e.g., 0-1 correctness).  
The expected performance of policy \(\pi\) in environment \(E\) is defined as
\begin{equation}
V_{E}(\pi)\triangleq \mathbb{E}_{(\tau,Y)\sim P^{\pi}_{E}}[\,U(\tau,Y)\,].
\end{equation}
The policy trained in environment \(L\) is denoted by \(\pi^\star\), satisfying
\begin{equation}
V_{L}(\pi^\star)\;\ge\;V_{L}(\pi),\qquad \text{for any feasible policy }\pi.
\end{equation}
The aim is to show that in environment \(L'\), the performance \(V_{L'}(\pi^\star)\) is not worse than other policies, and under common conditions strictly better.  
To establish this, first consider the relation between the distributional divergence of environments and performance difference.  
Since \(U\in[0,1]\), by the definition of total variation distance,
\begin{equation}
\begin{split}
\big|V_{L'}(\pi)-V_{L}(\pi)\big|
&= \Big|\,\mathbb{E}_{P^{\pi}_{L'}}[U]-\mathbb{E}_{P^{\pi}_{L}}[U]\,\Big| \\
&\le \mathrm{TV}\!\big(P^{\pi}_{L'},P^{\pi}_{L}\big).
\end{split}
\end{equation}
Assume there exists \(\varepsilon\in[0,1)\) such that for all classes \(y\) and actions \(a\),
\begin{equation}
\mathrm{TV}\!\big(K^{L'}_{a}(\cdot\mid y),\,K^{L}_{a}(\cdot\mid y)\big)\;\le\;\varepsilon.
\end{equation}
Then, for any policy \(\pi\) and horizon \(T\), by the chain inequality of product measures and the union bound, the trajectory distribution satisfies
\begin{equation}
\mathrm{TV}\!\big(P^{\pi}_{L'},P^{\pi}_{L}\big)\;\le\;1-(1-\varepsilon)^{T}\;\le\;T\,\varepsilon.
\end{equation}
Hence, the performance deviation is uniformly bounded as
\begin{equation}
\big|V_{L'}(\pi)-V_{L}(\pi)\big|\;\le\;T\,\varepsilon.
\end{equation}
Applying this to \(\pi^\star\) and any comparison policy \(\pi\) gives
\begin{equation}
V_{L'}(\pi^\star)\;\ge\;V_{L}(\pi^\star)-T\varepsilon,
V_{L'}(\pi)\;\le\;V_{L}(\pi)+T\varepsilon.
\end{equation}
Subtracting the two inequalities and using the optimality of \(\pi^\star\) in environment \(L\) yields
\begin{equation}
V_{L'}(\pi^\star)-V_{L'}(\pi)
\;\ge\; \big(V_{L}(\pi^\star)-V_{L}(\pi)\big)\;-\;2T\varepsilon.
\end{equation}
Whenever the performance gap in environment \(L\) satisfies
\begin{equation}
V_{L}(\pi^\star)-V_{L}(\pi)\;>\;2T\varepsilon,
\end{equation}
it follows that
\begin{equation}
V_{L'}(\pi^\star)\;>\;V_{L'}(\pi).
\end{equation}
This demonstrates that if the alternative environment \(L'\) is sufficiently close to the training environment \(L\) in terms of class-conditional observation kernels, and the margin in environment \(L\) exceeds the maximal penalty \(2T\varepsilon\), then the superiority of the trained agent transfers to the new environment.  

The result can also be expressed as a robust generalization inequality: for any family of environments
\begin{equation}
\mathcal{E}=\{E:\sup_{a,y}\mathrm{TV}(K^{E}_{a}(\cdot\mid y),K^{L}_{a}(\cdot\mid y))\le\varepsilon\},
\end{equation}
it holds that
\begin{equation}
\begin{split}
\inf_{E\in\mathcal{E}}\big(V_{E}(\pi^\star)-V_{E}(\pi)\big) \ge{} \\
\quad \big(V_{L}(\pi^\star)-V_{L}(\pi)\big) - 2T\varepsilon.
\end{split}
\end{equation}
The right-hand side equals the training-environment margin minus the worst-case shift penalty. For sufficiently small $\varepsilon$ (or a large margin), the bound remains positive.

Thus, when observation kernels differ only by small perturbations, the policy $\pi^\star$ trained in $L$ remains superior in any $L'$ with at most a linear penalty $T\varepsilon$, so an agent trained with one LLM environment also performs better with others.
\end{proof}

\begin{algorithm*}[t]
\caption{Prompt-R1: Collaborative Automatic Prompting Framework via End-to-end Reinforcement Learning}
\label{alg:promptr1}
\SetAlgoLined
\SetKwInOut{Require}{Require}
\SetKwInOut{Ensure}{Ensure}

\Require{Question $q$, small-scale LLM $S$, large-scale LLM $L$, policy $\pi_\theta$, reward function $R(\cdot)$, initial prompt template $a_{\text{tmpl}}$, maximum interaction turns $T$}
\Ensure{Final answer $y$}

1: \textbf{// Stage A: Collaborative Prompt Initialization} \\
2: Initialize: interaction history $H_0=[\,]$, response $r_{\text{prompt}}^0=\emptyset$ \\
3: Joint input: provide $(q, a_{\text{tmpl}})$ to $S$ \\
4: First think and prompt: $(a_{\text{think}}^1, a_{\text{prompt}}^1) \leftarrow S_{\pi_\theta}(q, a_{\text{tmpl}})$ \\
5: First response: $r_{\text{prompt}}^1 \sim L(\cdot \mid H_0, a_{\text{prompt}}^1)$ \\
6: First update history: $H_1 \leftarrow \{(a_{\text{prompt}}^1, r_{\text{prompt}}^1)\}$ \\

7: \textbf{// Stage B: Multi-turn Collaborative Prompt Interaction} \\
8: \textbf{for} $t=2$ \textbf{to} $T$ \textbf{do} \\
9: \quad Plan think: $a_{\text{think}}^t \leftarrow S_{\pi_\theta}(q, a_{\text{tmpl}}, H_{t-1})$ \\
10: \quad Generate prompt: $a_{\text{prompt}}^t \leftarrow S_{\pi_\theta}(q, a_{\text{tmpl}}, H_{t-1})$ \\
11: \quad LLM response: $r_{\text{prompt}}^t \sim L(\cdot \mid H_{t-1}, a_{\text{prompt}}^t)$ \\
12: \quad Update history: $H_t \leftarrow H_{t-1} \cup \{(a_{\text{prompt}}^t, r_{\text{prompt}}^t)\}$ \\
13: \textbf{end for} \\

14: Final think \& answer: $(a_{\text{think}}^T, y) \leftarrow S_{\pi_\theta}(q, a_{\text{tmpl}}, H_T)$ \\
15: Answer: $y = \arg\max_{y \in \mathcal{Y}^*} S_{\pi_\theta}(y \mid q, H_T)$ \\

16: \textbf{// Stage C: End-to-End Reinforcement Learning Optimization} \\
17: Sample trajectories: $\{\tau_i\}_{i=1}^N \sim \pi_\theta$ \\
18: \textbf{for each} $\tau_i$ \textbf{do} \\
19: \quad Compute reward: $R(\tau_i) = -k + R_{\text{format}}(\tau_i) + \mathbb{I}\{R_{\text{format}}(\tau_i) = k\} \cdot R_{\text{answer}}(\tau_i)$\\ 
20: \quad Compute advantage: $\hat{A}^{(\tau_i)} = \frac{R^{(\tau_i)} - \bar{R}}{\sqrt{\tfrac{1}{M}\sum_{j=1}^M (R^{(j)}-\bar{R})^2 + \varepsilon}}$ \\
21: \textbf{end for} \\

22: GRPO-based update policy: \\
23: $\displaystyle\mathcal{J}_{\text{GRPO}} \propto \sum_{i=1}^N \sum_{t=1}^{|\tau_i|} \min\left( \rho_\theta(w_t^{(i)}), \text{clip}(\rho_\theta(w_t^{(i)}), 1 \pm \epsilon) \right) \hat{A}(\tau_i)$ \\

24: \quad where $\displaystyle\rho_\theta(w_t^{(i)}) = \frac{\pi_\theta(w_t^{(i)} \mid \tau_{<t}^{(i)})}{\pi_{\theta_{\text{old}}}(w_t^{(i)} \mid \tau_{<t}^{(i)})}$ \\

25: Parameter update: $\theta \leftarrow \theta - \eta \nabla_\theta(-\mathcal{J}_{\text{GRPO}})$

\end{algorithm*}

\section{Prompt-R1 Algorithm Details}
\label{prompt-R1_algorithm}
Prompt-R1 is a multi-turn prompt interaction framework built on collaboration between a small-scale LLM and a large-scale LLM: the small-scale LLM handles planning, prompt generation, and answer output, while the large-scale LLM provides the corresponding responses for the prompts. The process is divided into three connected stages: first, given a question explanation, the small-scale LLM constructs a collaborative template with structural guidance and produces the first prompt, and the large-scale LLM returns an initial response to form first interaction history; then, during multi-turn interaction, the small-scale LLM continuously generates reasoning and the next prompt from the accumulated history, the large-scale LLM provides new responses and extends the history until the termination condition is reached, and the accumulated trajectory is utilized to generate the final answer; finally, end-to-end reinforcement learning is applied to update the prompt policy of the small-scale LLM. The reward function jointly evaluates format compliance and answer correctness, enabling the small-scale LLM to gradually learn how to effectively drive the large-scale LLM within a limited budget. This coordination reduces prompting waste and stabilizes interaction dynamics. This design enables progressive and adaptive reasoning, resulting in improved accuracy and stability on complex tasks.

\textit{Training and Inference Flow.}
During training, the algorithm proceeds in three stages (A→B→C): the small-scale LLM initializes the template and triggers the large-scale LLM for an initial response, then enters multi-turn interaction to generate prompts, receive responses, and terminate with an answer in a sequential manner, and finally updates the policy in stage C using rewards and advantages. During testing, it runs only two stages (A→B) in a simplified form: initialization and multi-turn interaction, after which the final reasoning and answer are produced directly without parameter updates.

\textit{Complexity Analysis.}
The computational complexity of Prompt-R1 mainly comes from initialization, multi-turn interaction, and reinforcement learning optimization. The initialization stage involves one call to the small-scale LLM and a call to the large-scale LLM, which is a constant overhead. The multi-turn interaction stage requires up to $T$ rounds in the worst case, where each round includes one planning step by the small-scale LLM and one call to the large-scale LLM, yielding time complexity $O(T)$. The memory consumption grows linearly with the history length, which can be controlled through windowing or summarization. The reinforcement learning stage requires sampling $N$ trajectories per update, each trajectory containing up to $T$ prompt-response steps, leading to complexity $O(NT)$. It also requires storing trajectory information for reward and advantage computation. In total, the complexity of Prompt-R1 is $O(NT+T)$, scaling linearly with the number of rounds and sampled trajectories during training, while inference requires $O(T)$. Since the small-scale LLM is responsible for planning and refining prompts, the large-scale LLM is more focused, ensuring stability on complex reasoning tasks.

\section{Dataset Details}
\label{dataset}
We selected 12 public datasets (including multi-hop reasoning, mathematical calculation, common sense question answering, and text generation) for training and testing. Eight of these datasets were used for training and testing. Four datasets were used for out-of-distribution testing to verify the generalization of the proposed Prompt-R1. These datasets are as follows:

\noindent
$\bullet$ \textbf{2WikiMultihopQA}: A multi-hop QA dataset requiring reasoning across two Wikipedia articles, designed to test information integration across documents. 

$\bullet$ \textbf{HotpotQA}: A large-scale multi-hop QA corpus with questions covering diverse topics, where answers typically demand linking multiple paragraphs. 

$\bullet$ \textbf{GSM8K}: A collection of grade school math word problems with concise statements, emphasizing step-by-step calculation and accurate numeric results. 

$\bullet$ \textbf{DAPO Math}: A dataset of algebraic and multi-step mathematical problems, often involving complex formula derivations and logical process reasoning. 

$\bullet$ \textbf{MusiQue}: A composite QA dataset built from multiple sub-questions, where answers require progressive reasoning across factual and thematic domains. 

$\bullet$ \textbf{PopQA}: A large-scale corpus of common knowledge and popular culture questions, featuring short queries, clear answers, and broad topical coverage. 

$\bullet$ \textbf{BookSum}: A long-text summarization dataset derived from novels and book chapters, providing multi-level alignments between extended texts and summaries. Since the input content of this dataset is too long, we used a LLM to construct structured data that is easy to train using the chapter field of the dataset. More details are in our training dataset.

$\bullet$ \textbf{WritingPrompts}: A creative writing corpus collected from community prompts and corresponding stories, encompassing diverse narrative styles and genres. 

$\bullet$ \textbf{MathQA}: A math QA dataset curated from multi-domain exam problems, covering arithmetic, algebra, geometry, probability, and other sub-disciplines. 


$\bullet$ \textbf{TriviaQA}: A knowledge-intensive dataset with questions from trivia websites and Wikipedia, containing a wide range of facts and lesser-known topics. 

$\bullet$ \textbf{XSum}: A dataset of BBC news articles paired with one-sentence extreme summaries, emphasizing highly concise abstraction of essential information. 

$\bullet$ \textbf{SQuAD v2}: A Wikipedia-based QA dataset combining answerable and unanswerable questions, constructed to evaluate comprehension under mixed conditions.

To ensure consistency and fairness for training and testing, 5,120 instances were randomly sampled from each of the eight selected datasets for training, resulting in a total of 40,960 training instances. To evaluate the generalization performance of Prompt-R1, 128 instances were randomly sampled from each of the four out-of-distribution and eight trained datasets.

\section{Baseline Details}
\label{baselines}
To accurately evaluate the performance of Prompt-R1, we conducted comparative experiments against multiple baselines. These baselines can be broadly divided into two categories: those based on GPT-4o-mini and those based on Qwen3-4B.

\subsection{Baselines with GPT-4o-mini}
$\bullet$ \textbf{GPT-4o-mini}~\citep{hurst2024gpt}: A lightweight variant of GPT-4o optimized for cost and latency, while retaining strong language and reasoning abilities. As a baseline, it is tested with standard instruction prompting without retrieval or tool use, measuring the model’s inherent generation capacity under constrained resources.

$\bullet$ \textbf{CoT (GPT-4o-mini)}\citep{wei2022chain,hurst2024gpt}: Chain-of-thought prompting applied to GPT-4o-mini, leveraging its stronger reasoning capacity to generate intermediate steps. It serves as a higher-capacity baseline for reasoning comparison against smaller backbones.  

$\bullet$ \textbf{GEPA}\citep{agrawal2025gepa}: A prompt optimizer that combines genetic search and Pareto frontier exploration with natural language reflection. It diagnoses trajectories (e.g., reasoning, tool calls) in natural language and proposes updates, achieving large quality gains with fewer rollouts compared to traditional RL-based methods.  

$\bullet$ \textbf{TextGrad}\citep{yuksekgonul2024textgrad}: A gradient-inspired optimization framework that treats natural language feedback as approximate gradients. It iteratively improves prompts or variables in computation graphs, enabling general-purpose adaptation across tasks such as QA, code, and molecule optimization without direct gradient access.  

$\bullet$ \textbf{OPRO}~\citep{yang2023large}: An optimization-by-prompting approach that leverages LLMs themselves as optimizers. At each step, the LLM generates and evaluates new candidate solutions described in natural language, iteratively refining prompts. It has been shown to outperform human-designed prompts on reasoning benchmarks.  

\subsection{Baselines with Qwen3-4B}
$\bullet$ \textbf{Qwen3-4B}~\citep{yang2025qwen3}: A 4B-parameter language model from Alibaba Cloud, serving as a compact backbone for generation tasks. It provides strong efficiency in low-resource settings while maintaining competitive performance on reasoning benchmarks.  

$\bullet$ \textbf{SFT (Qwen3-4B)}~\citep{ouyang2022training}: A supervised fine-tuning baseline built on Qwen3, trained with human-annotated data to improve instruction following and response accuracy. It evaluates how standard supervised adaptation enhances the raw backbone’s capabilities.  

$\bullet$ \textbf{CoT (Qwen3-4B)}~\citep{wei2022chain,yang2025qwen3}: Chain-of-thought prompting applied to Qwen3, encouraging the model to generate step-by-step reasoning before producing the final answer. This improves logical consistency and performance on reasoning-intensive tasks.  

$\bullet$ \textbf{GRPO (Qwen3-4B)}\citep{guo2025deepseek}: Group Relative Policy Optimization, a reinforcement learning algorithm that normalizes rewards within sampled groups of trajectories. This reduces variance in policy updates, stabilizes training, and improves convergence efficiency compared to standard (Proximal Policy Optimization) PPO.  

\subsection{Other LLM Baselines }
$\bullet$ \textbf{Deepseek-V3}: Deepseek-V3 adopts advanced context understanding algorithms, optimized specifically for real-time conversation systems, supporting efficient multi-turn interactions. The model dynamically adjusts response strategies based on conversation history, providing a more intelligent user experience. It excels at maintaining context over extended interactions, ensuring the conversation flows naturally and remains coherent.

$\bullet$ \textbf{Grok-4-Fast}: Grok-4-Fast incorporates an efficient inference optimization mechanism, capable of handling complex natural language tasks with minimal latency. By combining knowledge distillation and optimization algorithms, the model significantly improves inference speed while maintaining high accuracy, making it suitable for real-time applications. Its speed makes it highly effective in time-sensitive scenarios, such as live customer support or real-time data processing.

$\bullet$ \textbf{Llama-4-Maverick}: Llama-4-Maverick leverages the latest multimodal learning techniques, integrating different data sources such as text and vision to handle more complex tasks. With its hierarchical semantic understanding framework, the model shows greater adaptability and accuracy across various domain-specific tasks. This allows it to be applied in a variety of fields, from healthcare to autonomous systems, where cross-modal understanding is crucial.

$\bullet$ \textbf{GPT-5}: GPT-5 achieves breakthroughs in natural language understanding and generation, particularly in long-text reasoning and cross-domain knowledge integration. The model utilizes a deep learning architecture and reinforcement learning optimization algorithms, enabling it to handle complex reasoning and generation tasks with greater precision. Its ability to integrate knowledge from different domains allows it to excel at complex problem-solving tasks, making it versatile across industries.

$\bullet$ \textbf{Gemini-2.5-Flash}: Gemini-2.5-Flash combines a fast inference engine and distributed computing technology, demonstrating exceptional efficiency in processing large-scale data. The optimized memory management and computational architecture ensure low latency and high throughput when handling massive datasets. This makes it an ideal choice for applications that require rapid data analysis, such as financial modeling or real-time sensor data processing.

$\bullet$ \textbf{Qwen-Plus}: Qwen-Plus incorporates an adaptive learning mechanism, allowing the model to continuously optimize its performance in dynamic environments. Combining multi-task learning and self-supervised learning, it provides efficient solutions across a wide range of complex applications. Its ability to adapt to evolving data patterns makes it highly effective in dynamic environments, such as personalized recommendation systems or real-time analytics.

$\bullet$ \textbf{GPT-OSS-20B}: GPT-OSS-20B is an open-source 20B-parameter LLM optimized for high-performance inference. By integrating knowledge graphs and cross-modal reasoning, it excels across a range of tasks. Its flexibility enables seamless customization for diverse applications, ranging from research to enterprise solutions.

\begin{table*}[t]
\vskip 0.15in
\begin{center}
\fontsize{8.6pt}{8.6pt}\selectfont
\setlength{\tabcolsep}{36pt}
\renewcommand{\arraystretch}{1.0}
\begin{tabular}{l l l}
\toprule
\textbf{Method} & \textbf{Hyperparameter} &  \textbf{Value}\\
\midrule
\multirow{8}{*}{\textbf{Prompt-R1 (Agent)}} 
 & Batch size & 128 \\
 & Learning Rate & $1 \times 10^{-6}$ \\
 & PPO Mini-batch Size & 64 \\
 & Micro-batch Size per GPU & 2 \\
 & Policy Repeat Count & 5 \\
 & Maximum Context Length & 8192 \\
 & Maximum Response Length & 8192 \\
 & Maximum Single-turn Response Length & 1024 \\
 & Maximum Turns & 5 \\
 & Training Epochs & 1 \\
\bottomrule
\end{tabular}
\end{center}
\vskip -0.15in
\caption{Training hyperparameter settings for Prompt-R1 agent (Qwen3-4B).}
\label{tab: impl}
\end{table*}

\begin{table*}[t]
\vskip -0.1in
\begin{center}
\fontsize{8.6pt}{8.6pt}\selectfont
\setlength{\tabcolsep}{43pt}
\renewcommand{\arraystretch}{1.0}
\begin{tabular}{l l l}
\toprule
\textbf{Method} & \textbf{Hyperparameter} &  \textbf{Value}\\
\midrule
\multirow{8}{*}{\textbf{Qwen3-4B (SFT)}} 
 & Finetuning Type & Lora \\
 & Lora Rank & 8 \\
 & Batch Size & 4 \\
 & Gradient Accumulation Steps & 8 \\
 & Learning Rate & $1 \times 10^{-4}$ \\
 & Epochs & 3.0 \\
 & Learning Rate Scheduler Type & Cosine \\
 & Warmup Ratio & 0.1 \\
 & BF16 & True \\
\bottomrule
\end{tabular}
\end{center}
\vskip -0.15in
\caption{Training hyperparameter settings of SFT for Qwen3-4B.}
\label{tab:sft_qwen}
\vskip -0.15in
\end{table*}

\section{Evaluation Details}
\label{evaluation_metrics}

\textbf{Exact Match.} 
The Exact Match (EM) metric measures whether the model's predicted answer exactly matches the ground truth. If the predicted answer matches the ground truth exactly, the EM value for that sample is 1; otherwise, it is 0. The formula for EM is as follows:
\begin{equation}
\text{EM} = \frac{1}{N} \sum_{i=1}^{N} \mathbb{I}(\text{norm}(y_i) = \text{norm}(y_i^*))
\end{equation}
where \( y_i \) is the predicted answer, \( y_i^* \) is the ground truth, \( \mathbb{I} \) is the indicator function, \( \text{norm}(\cdot) \) is the normalization function, and \( N \) is the total number of samples.


\textbf{F1-score}.
The F1-score (F1) metric measures the overlap between the predicted answer and the ground truth answer, considering both Precision and Recall. The formula for F1-score is as follows:
\begin{equation}
\text{F1} = \frac{2 \cdot \text{Precision} \cdot \text{Recall}}{\text{Precision} + \text{Recall}}
\end{equation}
where:
\begin{equation}
\text{Precision} = \frac{| \text{tokens}(y_i) \cap \text{tokens}(y_i^*) |}{|\text{tokens}(y_i)|}
\end{equation}
and
\begin{equation}
\text{Recall} = \frac{| \text{tokens}(y_i) \cap \text{tokens}(y_i^*) |}{|\text{tokens}(y_i^*)|}.
\end{equation}

\textbf{Semantic Similarity}.
The Semantic Similarity (SSim) metric measures the semantic similarity between the predicted answer and the ground truth answer, typically computed using cosine similarity. The formula for SSim is as follows:
\begin{equation}
\text{SSim} = \frac{1}{N} \sum_{i=1}^{N} \cos(\text{Enc}(k_i^{\text{pred}}), \text{Enc}(k_i^{\text{gold}}))
\end{equation}
where \( \text{Enc}(\cdot) \) represents the semantic encoding function, \( k_i^{\text{pred}} \) and \( k_i^{\text{gold}} \) represent the semantic representations of the predicted answer and the ground truth, respectively, and the cosine similarity between them is computed.

\begin{figure*}[t]
\centering
\includegraphics[width=1.0\textwidth]{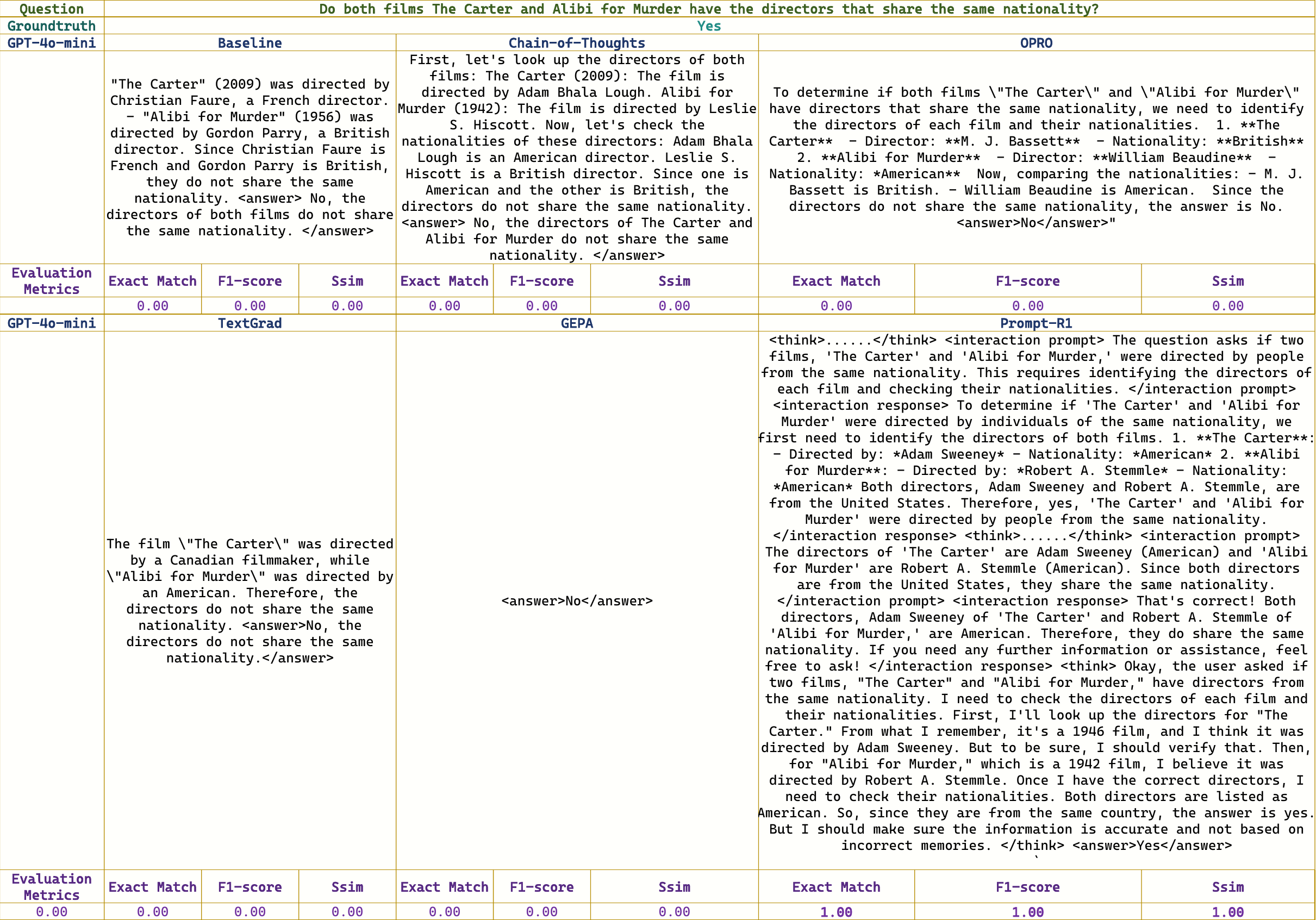} 
\caption{Case studies of prompt optimization methods, including Baseline (GPT-4o-mini), Chain-of-Thoughts (GPT-4o-mini), automatic prompting optimization methods (OPRO, TextGrad, and GEPA), and Prompt-R1.}
\label{fig:prompt-r1-case_study}
\end{figure*}

\section{Implementation Details}
\label{implementation_metrics}
To ensure reproducibility and fair comparison, we summarize the hyperparameter configurations for Prompt-R1 in Table~\ref{tab: impl}. This table reports hyperparameters of the Prompt-R1 agent, including batch size, learning rate, PPO mini-batch size, micro-batch size per GPU, policy repeat count, maximum context length, maximum response length, maximum single-turn response length, maximum turns, and training epochs. For training the Prompt-R1 agent, all our experiments are conducted on a GPU server with eight NVIDIA A100 GPUs. Each GPU has a memory capacity of 40 GB. For the GPT-OSS-20B as the environment, we adopted a local deployment method, deploying it on another server equipped with eight Nvidia A100 GPUs for Prompt-R1 agent training, to achieve zero-cost training of the Prompt-R1 agent. In the double-constrained reward, the $\alpha$ is 0.4; the $\beta$ is 0.25; the $\gamma$ is 0.25; the $\delta$ is 0.1; and the $k$ is 1.0.


For the experiments based on Qwen3-4B, we used the same prompt to constrain the answer format and LLaMA-Factory for training and evaluation on the same GPU server as Prompt-R1. The hyperparameter settings of the GRPO for Qwen3-4B are the same as Prompt-R1. We conducted multiple experiments and selected the optimal parameters of the SFT for Qwen3-4B. The hyperparameters of the SFT for Qwen3-4B are illustrated in Table \ref{tab:sft_qwen}.

For the automatic prompting optimization (including OPRO, TextGrad, and GEPA), which is based on GPT-4o-mini, and the LLM baselines (e.g., Deepseek-V3, Grok-4-fast, LLaMA-4-Maverick, GPT-5, Gemini-2.5-flash, and Qwen-Plus), we also used the same prompt to format the answers and call the official APIs for evaluation. For a fair comparison, we conducted three independent runs under identical settings for both Prompt-R1 and all baselines and reported the averaged results. Regarding the extremely poor performance of Gemini-2.5-flash on the DAPO Math dataset, as shown in Figure \ref{fig:6-models-8-datasets-radar}, we examined the content returned after calling the API and identified two main reasons for this. First, it faces a performance bottleneck in mathematical computational capabilities; second, it fails to follow prompt instructions as accurately as other LLMs in providing correctly formatted answers. 

\section{Case Study}
\label{case_study}
As shown in Figure \ref{fig:prompt-r1-case_study}, a concrete case study is presented to evaluate various baselines and the Prompt-R1 in a realistic reasoning scenario. The task requires determining whether the directors of The Carter and Alibi for Murder share the same nationality, with the ground-truth answer being “Yes”. Results indicate that all baseline models, including GPT-4o-mini, Chain-of-Thoughts, OPRO, TextGrad, and GEPA, failed to produce the correct response. They incorrectly associated The Carter with unrelated directors (e.g., Christian Faure, M. J. Bassett, Adam Bhala Lough) and linked Alibi for Murder to directors of mismatched nationalities, reflecting entity-linking drift and instability in factual recall. In contrast, Prompt-R1 correctly identified the directors of both films and determined their shared nationality, yielding the accurate answer. Its reasoning trajectory exhibits a stronger structural pattern: the model first decomposes the query into two atomic subtasks (identifying each director) and then aggregates their nationality attributes for logical comparison, thereby achieving consistent cross-entity factual verification. This case clearly demonstrates that the multi-turn, prompt-interactive reasoning mechanism of Prompt-R1 substantially enhances the LLM’s stability in recognizing cross-entity relations (e.g., nationality, profession, affiliation) and achieves superior factual alignment and reasoning robustness.

\section{Limitations}
\label{Limitations}

Despite its strong empirical performance across various tasks, Prompt-R1 has several limitations. First, its heavy reliance on historical context is a key structural constraint. As multi-turn interactions progress, the quality of initial prompts and responses becomes critical for sustaining accurate and coherent reasoning. Even subtle errors or ambiguities introduced at early stages may accumulate rapidly via error propagation, affecting the reliability and accuracy of the final output. Consequently, when the historical context is incomplete or noisy, its reasoning ability can be compromised. Additionally, the method depends heavily on continuous and dynamic updates to the historical context. If these updates fail to capture new, relevant information promptly, it can lead to substantial information loss or incorrect decisions, thereby limiting the model's practical flexibility and effectiveness.

\section{Future Work}
\label{Future Work}
To further enhance the overall performance and robustness of Prompt-R1, future work should focus on comprehensively improving both scalability and efficiency. Specifically, optimizing the complex multi-turn prompt interaction process through techniques such as context compression or selective summarization could significantly reduce computational overhead while improving inference speed and accuracy. Additionally, refining the reinforcement learning component, particularly the design of the reward mechanism, can further boost learning efficiency and dynamic adaptability. To expand applicability across diverse domains, incorporating domain adaptation and transfer learning strategies could strengthen its ability to handle heterogeneous cross-domain tasks and generalize effectively to unseen scenarios. Addressing long-range dependencies in multi-turn reasoning may be achieved by exploring advanced memory mechanisms, ensuring contextual coherence over extended interaction histories. Lastly, optimizing reward functions and incorporating diverse feedback would enhance performance on increasingly complex tasks.

\section{Applicability Analysis}
\label{Applicability Analysis}
Prompt-R1, with its advanced multi-turn reasoning and dynamic context updating capabilities, demonstrates significant potential for application in highly knowledge-intensive domains that require rigorous logical deduction. Particularly in critical fields such as law, healthcare, and finance, Prompt-R1 can leverage powerful large language models to handle increasingly complex reasoning tasks efficiently while ensuring data privacy even in resource-constrained environments. Moreover, by integrating reinforcement learning techniques, Prompt-R1 is not only capable of handling traditional supervised tasks but also adapts seamlessly to complex dynamic environments by continuously optimizing its reasoning strategies via intrinsic self-correction mechanisms, thus greatly enhancing the system's adaptive intelligence and long-term planning proficiency. Overall, Prompt-R1 provides strong support for trustworthy, transparent, and intelligent decision-making in knowledge-intensive fields, offering promising applications in a wide range of challenging real-world domains with its robust reasoning capabilities and adaptability.
\end{document}